%% file: main.tex
\documentclass{article}

\usepackage[main, final]{neurips_2026}

\usepackage[utf8]{inputenc} %
\usepackage[T1]{fontenc}    %
\usepackage{hyperref}       %
\hypersetup{pdftitle={MICE: Multi-animal Interaction Context Encoder -- A Hierarchical Foundation Model for Mouse Behavior},pdfauthor={Yen-Shuo Su, Weihan Li, Anqi Wu}}
\usepackage{url}            %
\usepackage{booktabs}       %
\usepackage{amsfonts}       %
\usepackage{nicefrac}       %
\usepackage{microtype}      %
\usepackage{xcolor}         %
\usepackage{amsmath}
\usepackage{graphicx}
\usepackage{multirow}
\usepackage{bm}
\usepackage{diagbox}

\title{MICE: Multi-animal Interaction Context Encoder ---\texorpdfstring{\\}{ }A Hierarchical Foundation Model for Mouse Behavior}

\author{%
  Yen-Shuo Su \qquad Weihan Li \qquad Anqi Wu \\
  Georgia Institute of Technology \\
  Atlanta, GA, USA \\
  \texttt{\{ericsu, weihanli, anqiwu\}@gatech.edu} \\
}

\begin{document}

\maketitle

\begin{abstract}
Quantifying mouse behavior from pose is a foundational tool in neuroscience, ethology, drug discovery, and animal welfare. Self-supervised foundation models are a natural fit, but a foundation model for mouse behavior must handle social behavior --- many behaviors that matter (aggression, courtship, mounting, allogrooming) are inherently multi-animal. Existing self-supervised pose models fall short: they encode each animal independently and discard interaction signal, hard-code the animal count, and pretrain one dataset at a time. We introduce \textbf{MICE} (Multi-animal Interaction Context Encoder), a hierarchical foundation model for mouse behavior that pretrains jointly on six mouse-pose corpora spanning one to four mice and 7 to 27 keypoints. MICE combines a \emph{hierarchical individual encoder} capturing per-mouse kinematics at multiple temporal scales with a \emph{Perceiver-style social encoder} whose learned latent codebook decouples parameter count from animal count, so a single checkpoint serves any group size without retraining. Across four evaluation protocols spanning six datasets, MICE matches or exceeds prior keypoint-based baselines, and layer-wise probing shows the social stage incrementally improves discriminability beyond the individual encoder. Code, weights, and data are available at \url{https://github.com/eric0708/MICE}.
\end{abstract}

\section{Introduction}

Automated quantification of mouse behavior from video is increasingly central to neuroscience, ethology, drug discovery, and animal welfare~\citep{anderson2014toward,datta2019computational,pereira2020quantifying}, with pose-estimation toolkits~\citep{mathis2018deeplabcut,pereira2022sleap,segalin2021mouse} now producing keypoint trajectories at scale. The bottleneck has shifted from \emph{tracking} to \emph{interpretation}: turning $(x,y)$ keypoint sequences into ethologically meaningful annotations. Self-supervised foundation models are a natural way to bridge this gap, but a foundation model for mouse behavior must handle more than per-animal kinematics --- many of the most important behaviors (aggression, courtship, mounting, allogrooming) are intrinsically \emph{social}. Existing self-supervised pose models~\citep{sun2021task,azabou2023relax,stoffl2024elucidating} were not designed to scale into a foundation-model regime, for three reasons. \textbf{(1)~Pooling destroys interaction structure}~\citep{azabou2023relax,qi2017pointnet}: per-animal representations collapsed via mean- or max-pooling lose the relational signal defining dyadic behaviors. \textbf{(2)~Fixed-ordering or fixed-$N$ inputs}~\citep{sun2021task,stoffl2024elucidating}: concatenating multi-animal keypoints hard-codes both the animal count $N$ and the ordering at training time. \textbf{(3)~Skeletal heterogeneity}: different labs annotate different keypoint sets, so a model trained on one skeleton cannot accept input from another without explicit alignment.

We introduce \textbf{MICE} (Multi-animal Interaction Context Encoder), a hierarchical self-supervised foundation model for mouse behavior that pretrains jointly across heterogeneous mouse-pose datasets with variable animal counts. The architecture makes no mouse-specific assumptions and extends to other tracked species; we focus on mice because they are the dominant model organism in neuroscience and the only species with the public keypoint datasets and ethologically-validated labels needed for rigorous evaluation. Our contributions are: (i)~a \textbf{mouse-behavior foundation model} pretrained jointly across six corpora --- four published datasets (MABe22~\citep{sun2023mabe22}, CalMS21~\citep{sun2021multi}, CRIM13~\citep{burgos2012social}, MouseLockbox~\citep{reiske2025mouse}) and two codenamed mouse tracks from the MABe 2025 challenge~\citep{mabe2025} (AdaptableSnail, NiftyGoldfinch) --- aligned through a universal 12-keypoint vocabulary, a regime no prior keypoint-based behavior model has demonstrated; (ii)~a \textbf{hierarchical individual encoder} with widening attention windows, capturing per-mouse kinematics at multiple temporal scales; (iii)~a \textbf{Perceiver-style social encoder} that processes an arbitrary number of mice through a learned latent codebook, letting a single checkpoint serve any group size without retraining; and (iv)~a \textbf{comprehensive evaluation} across all six datasets and four evaluation protocols, where MICE matches or exceeds prior keypoint-based baselines, with layer-wise probing showing that the social stage incrementally improves discriminability beyond the individual encoder, plus controlled tests of what the social stage contributes (\S\ref{sec:mechanism}).

\begin{figure}[t]
\centering
\includegraphics[width=\linewidth]{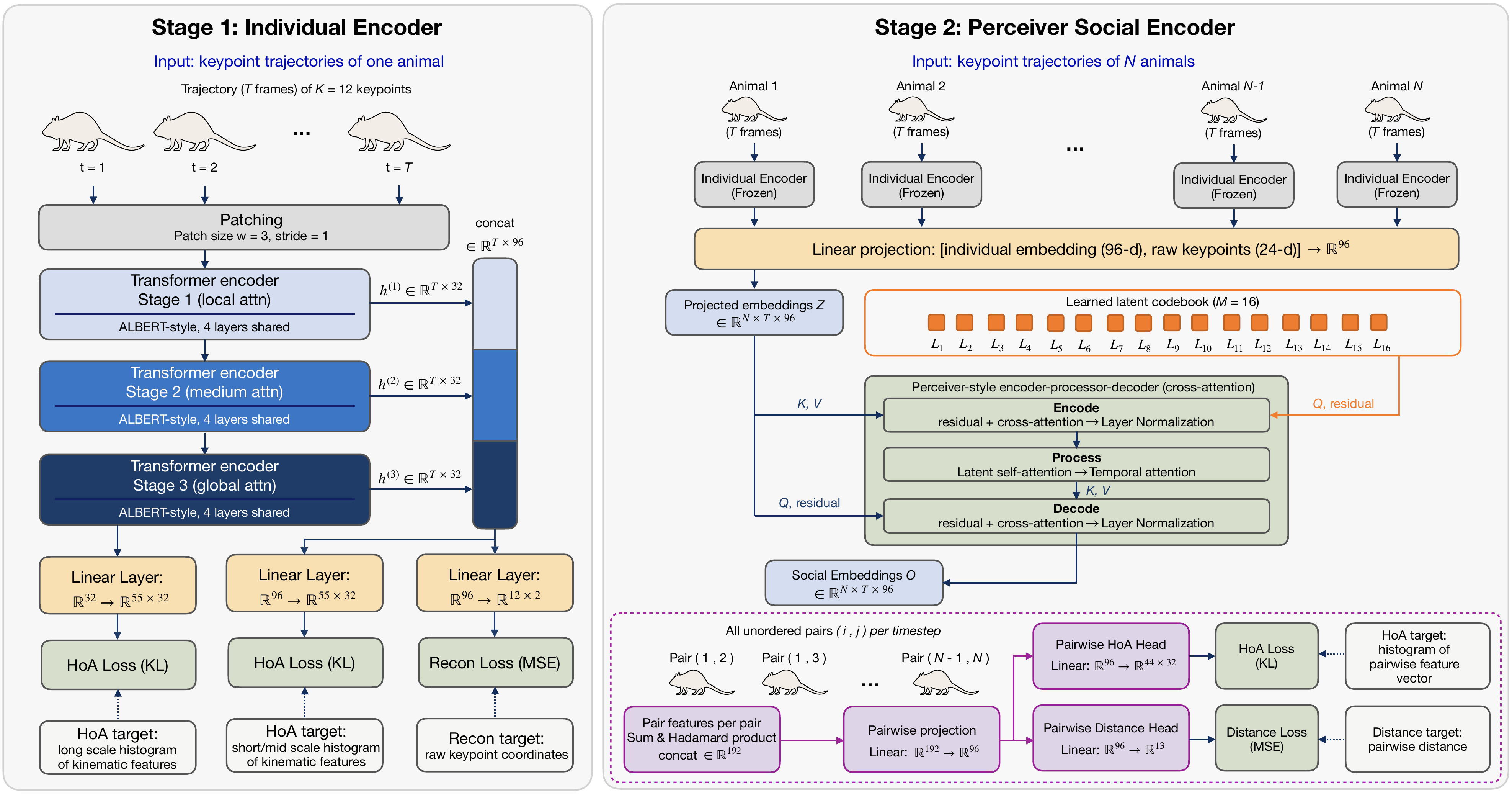}
\caption{The MICE architecture. \textbf{Stage 1 (left, individual encoder):} a per-animal pose sequence is tokenized by a patching layer, then processed through three ALBERT-style transformer stages of increasing attention scope (local, medium, global) with weights shared within each stage. Per-stage outputs are concatenated and supervised by three losses: a long-scale Histogram-of-Activities (HoA) loss on Stage 3's output, a short/mid-scale HoA loss on the concatenated output, and a span-masked keypoint reconstruction loss. \textbf{Stage 2 (right, social encoder):} the frozen individual encoder is applied per mouse; per-animal embeddings are fused with raw keypoints and passed to a Perceiver with a learned latent codebook. The latents cross-attend over animals (encode), refine across time (process), and are read back to each animal (decode), producing per-animal social embeddings. Animal pairs are formed by an order-invariant projection and supervised by pairwise HoA and pairwise distance losses.}
\label{fig:arch}
\end{figure}

\section{Related Work}

\paragraph{Pose-based behavior analysis.} Classical pipelines hand-engineer features (velocity, distance, angles) for Random Forests or HMMs~\citep{kabra2013jaaba,hong2015automated}. Deep approaches~\citep{nilsson2020simple,segalin2021mouse} learn from keypoints with supervised CNNs/RNNs but require large labeled corpora.

\paragraph{Self-supervised pose representation.} BAMS~\citep{azabou2023relax} predicts a histogram of binned action features (HoA) from a dual-branch dilated TCN, the closest precursor to our individual-stage objective. TBERT~\citep{sun2023mabe22} applies BERT-style masked modeling to keypoints; hBehaveMAE~\citep{stoffl2024elucidating} introduces a hierarchical masked autoencoder over (time$\times$mice$\times$feature) tensors but fixes both the skeleton and the animal count; PointNet~\citep{qi2017pointnet} treats keypoints as an unordered point cloud. Unlike these, MICE's two-stage approach explicitly separates intra-animal kinematics from inter-animal interaction.

\paragraph{Video foundation models for mouse behavior.} A parallel line of work pursues behavioral foundation models from raw video. BEAST~\citep{wang2026animal} uses masked autoencoding plus temporal contrastive learning; \citet{sun2024video} apply VideoPrism~\citep{zhao2024videoprism} to mouse behavior datasets; generative video models have also been used to relate neural dynamics to behavior~\citep{wang2024exploring}. Video preserves appearance but is lab-specific (lighting, camera, arena), while keypoints abstract these confounds and are far cheaper at training and inference. MICE focuses on the keypoint regime, where this abstraction enables transfer across labs with different recording conditions.

\paragraph{Social interaction modeling and foundation models for behavior.} Existing social models hard-code animal count $N$~\citep{eyjolfsdottir2016learning,sun2023mabe22} or rely on $\mathcal{O}(N^2)$ GNN attention~\citep{kipf2018neural}. The Perceiver~\citep{jaegle2021perceiver} and Set Transformer~\citep{lee2019set} use learned latent queries to handle variable-length input; MICE is, to our knowledge, the first application of Perceiver-style bottlenecks to multi-animal social behavior. Foundation models exist for DNA~\citep{nguyen2023hyenadna}, proteins~\citep{lin2023evolutionary}, and brain activity~\citep{azabou2023unified}. MABe22~\citep{sun2023mabe22} and MABe 2025~\citep{mabe2025} introduced benchmarks with baselines evaluated on their own datasets but not across labs. MICE is the first behavior foundation model to pretrain jointly across multiple mouse-pose datasets with variable group sizes.

\section{Method}
\label{sec:method}

MICE has two stages (Figure~\ref{fig:arch}): an \emph{individual encoder} processes each mouse independently for per-animal kinematic representations at multiple temporal scales, and a \emph{social encoder} operates on these to model inter-animal interactions. Each stage is pretrained with self-supervision matched to its structure --- kinematics and dyadic relations respectively. We describe the universal keypoint vocabulary (\S\ref{sec:universal}), then each stage (\S\ref{sec:ind}, \S\ref{sec:soc}).

\subsection{Universal Mouse-Keypoint Space}
\label{sec:universal}

Joint pretraining across multiple datasets requires a shared input vocabulary, but different labs annotate different keypoint sets. We define a universal mouse-pose vocabulary of $K_{\max}{=}12$ keypoints following the MABe22 layout (nose, left/right ear, neck, left/right forepaw, center back, left/right hindpaw, tail base, tail middle, tail tip). Each dataset's native skeleton is mapped to this space by named-synonym matching, with missing keypoints zero-padded and masked from any feature that depends on them. Zero-padding is benign: keypoints have anatomical correspondence across labs, and the per-animal flat dimension $K_{\max}{\cdot}2{=}24$ is small enough that even CalMS21's 7-keypoint skeleton loses no more than half its slots to padding. This mapping is used only for combined pretraining; single-dataset models use each dataset's native keypoints.

\paragraph{Variable-$N$ handling.} The animal count is qualitatively different from keypoint count. Mouse anatomy bounds $K_{\max}{=}12$, but $N_{\max}$ has no such ceiling: group living, colony recordings, or cross-species extensions can push it beyond the dyadic regime, and a model trained at $N_{\max}{=}4$ cannot run on $N{=}6$ without retraining. MICE's Perceiver social encoder (\S\ref{sec:soc}) cross-attends over whichever animals are present, so a single checkpoint runs on any $N$. Fixed-input baselines (TBERT~\citep{sun2023mabe22}, hBehaveMAE~\citep{stoffl2024elucidating}) commit to $N_{\max}$ at training time; pooling-based baselines (BAMS~\citep{azabou2023relax}, PointNet~\citep{qi2017pointnet}) handle variable $N$ but lose interaction signal.

\subsection{Individual Encoder: Hierarchical Transformer}
\label{sec:ind}

The individual encoder produces per-mouse kinematic representations (Figure~\ref{fig:arch}, left). It takes a per-animal pose sequence ($T$ frames of $K{=}12$ keypoints in 2D) and returns a per-token feature vector at multiple temporal scales. Self-attention is bidirectional throughout (adapted from ALBERT~\citep{lan2019albert}): behavior analysis is post-hoc, so each token attends to past and future frames within its window.

\paragraph{Hierarchical design with continuous window schedule.} Mouse behavior unfolds at very different temporal scales: a paw step lasts tens of milliseconds, a sniff bout about a second, an attack episode several seconds. Prior designs commit to a small set of scales (hBehaveMAE~\citep{stoffl2024elucidating} uses discrete pooling levels; BAMS~\citep{azabou2023relax} uses a short- and long-term branch). Our design instead lets the attention scope grow continuously with depth (readout and supervision still use a few scales, Eqs.~2--3): across the 12 transformer layers ($i\in\{0,\ldots,11\}$), the attention window $\omega(i)$ grows linearly,
\begin{equation}
\omega(i) \;=\; \left\lfloor\,\omega_{\text{short}} + \frac{i}{11}\,\big(\omega_{\text{long}} - \omega_{\text{short}}\big)\,\right\rfloor,
\end{equation}
with $\omega_{\text{short}}{=}3$ and $\omega_{\text{long}}{=}T$ tokens, then rounded to the nearest odd integer so windows are symmetric around each query token. The full set of layers covers a continuum from short to long-range.

\paragraph{Stage organization.} The 12 layers are grouped into three stages of 4, indexed $g\in\{1,2,3\}$, for two reasons. First, it enables ALBERT-style weight sharing~\citep{lan2019albert}: layers within a stage share parameters, striking a middle ground between full sharing (one block handling every scale) and 12 independent layers (quadrupling the parameter count). Second, the stages define \emph{readout points} from which we extract one feature map per stage and concatenate them (Figure~\ref{fig:arch}, ``concat''):
\begin{equation}
\mathbf{H}_{\text{ind}} \;=\; \mathrm{concat}\!\left[\mathbf{h}^{(1)},\, \mathbf{h}^{(2)},\, \mathbf{h}^{(3)}\right] \;\in\; \mathbb{R}^{T \times 96}, \qquad D_{\text{ind}} = 3D = 96,
\end{equation}
where $\mathbf{h}^{(g)} \in \mathbb{R}^{T \times D}$ is stage $g$'s last-layer output and $D{=}32$. Three readouts are the minimum that gives a short/mid/long-range summary --- a single readout would force one representation to capture every scale, while per-layer probing would inflate the downstream parameter cost.

\paragraph{Self-supervised objectives.} The encoder is pretrained with three losses. Two are \emph{Histogram-of-Activities (HoA) prediction} losses, inspired by BAMS~\citep{azabou2023relax}: per-token histograms of a kinematic feature vector are predicted at two readout heads, yielding a short/mid-scale loss $\mathcal{L}_{\text{HoA}}^{\text{short/mid}}$ on $\mathbf{H}_{\text{ind}}$ and a long-scale loss $\mathcal{L}_{\text{HoA}}^{\text{long}}$ on $\mathbf{h}^{(3)}$ alone. The third is \emph{span-masked keypoint reconstruction} $\mathcal{L}_{\text{recon}}$ on $\mathbf{H}_{\text{ind}}$, anchoring the encoder to raw $(x,y)$ positions that HoA alone could leave unconstrained. The total objective combines all three:
\begin{equation}
\mathcal{L}_{\text{ind}} \;=\; \mathcal{L}_{\text{HoA}}^{\text{short/mid}} + \mathcal{L}_{\text{HoA}}^{\text{long}} + \mathcal{L}_{\text{recon}},
\end{equation}
where each $\mathcal{L}_{\text{HoA}}$ is a per-token KL divergence between predicted and target histograms and $\mathcal{L}_{\text{recon}}$ is the per-token MSE. Loss equations and target-window sizes are in Appendix~\ref{app:ind-losses}, and feature definitions in Appendix~\ref{app:features}.

\subsection{Social Encoder: Perceiver-based Model}
\label{sec:soc}

The social encoder operates on top of the per-animal representations from \S\ref{sec:ind} and produces an interaction-aware embedding for each mouse (Figure~\ref{fig:arch}, right). It runs on multi-animal pose with $N$ animals across $T$ frames.

\paragraph{Why a Perceiver.} Two requirements shape the design: handle a variable number of mice $N$ so a single trained model serves datasets with different group sizes, and preserve per-animal identity (pooling all animals into one vector would lose relational structure). A natural design satisfying both is direct $N$-to-$N$ cross-attention between animals, but its outputs are still built on animal representations --- the architecture cannot learn codes that specialize to relationships between animals (chasing, face-to-face confrontation, parallel locomotion). We instead use a Perceiver~\citep{jaegle2021perceiver}: a small set of learned latent codes ($M{=}16$ slots, the orange row in Figure~\ref{fig:arch}) reads from the $N$ animals, gets refined over time, and is passed back to each animal. Each latent is free to specialize to a behavioral motif, and at decode time each animal looks up which motifs it participates in.

\paragraph{Inputs.} The frozen individual encoder is applied per mouse, yielding hidden representations $\mathbf{H}_{\text{ind}}^{(n)} \in \mathbb{R}^{T\times 96}$. A linear projection (Figure~\ref{fig:arch}, ``Linear projection'') fuses each per-animal representation with raw keypoints into a per-token embedding $\mathbf{Z}_{n,t}\in\mathbb{R}^{96}$. During pretraining, contiguous temporal spans of $\mathbf{Z}$ are replaced with a mask token. Details in Appendix~\ref{app:soc-equations}.

\paragraph{Encode--process--decode.} The Perceiver applies three operations per timestep (Figure~\ref{fig:arch}, green block). The learned latent codebook $\mathbf{L}_0\in\mathbb{R}^{M\times 96}$ has $M{=}16$ slots; $\mathbf{Z}_{:,t}$ stacks the $N$ animal embeddings at timestep $t$. Using multi-head cross-attention $\mathrm{CrossAttn}(\mathbf{Q},\mathbf{K},\mathbf{V})$, latent self-attention $\mathrm{SelfAttn}$ (within a timestep), temporal self-attention $\mathrm{TempAttn}$ (across $T$ time tokens, per latent slot), and layer normalization $\mathrm{LayerNorm}$:
\begin{align}
\textbf{Encode:}\quad \mathbf{L}_t' &= \mathrm{LayerNorm}\!\Big(\mathbf{L}_0 + \mathrm{CrossAttn}\big(\mathbf{Q}{=}\mathbf{L}_0,\, \mathbf{K}{=}\mathbf{V}{=}\mathbf{Z}_{:,t}\big)\Big), \label{eq:encode}\\[3pt]
\textbf{Process:}\quad \mathbf{L}_t'' &= \mathrm{TempAttn}\!\Big(\mathrm{SelfAttn}\!\big(\mathbf{L}_t'\big)\Big), \label{eq:process}\\[3pt]
\textbf{Decode:}\quad \mathbf{O}_{n,t} &= \mathrm{LayerNorm}\!\Big(\mathbf{Z}_{n,t} + \mathrm{CrossAttn}\big(\mathbf{Q}{=}\mathbf{Z}_{n,t},\, \mathbf{K}{=}\mathbf{V}{=}\mathbf{L}_t''\big)\Big). \label{eq:decode}
\end{align}
Encode reads animal information into the latents; Process refines them within timestep then across time; Decode reads the refined latents back to each animal. $\mathbf{O}_{n,t}\in\mathbb{R}^{96}$ is the per-animal, per-timestep social embedding.

\paragraph{Pairwise projection.} Since social self-supervision is defined over pairs, we project each animal pair's social embeddings into a single representation. For each unordered pair $(i,j)$ with both animals valid at timestep $t$ (Figure~\ref{fig:arch}, dashed pink box):
\begin{equation}
\mathbf{p}_{ij,t} \;=\; \mathrm{Linear}\!\left([\,\mathbf{O}_{i,t} + \mathbf{O}_{j,t};\; \mathbf{O}_{i,t} \odot \mathbf{O}_{j,t}\,]\right) \;\in\; \mathbb{R}^{96},
\end{equation}
where $\mathrm{Linear}: \mathbb{R}^{192} \to \mathbb{R}^{96}$ applies to the concatenation of the elementwise sum and Hadamard product. Both operations are swap-invariant, so $\mathbf{p}_{ij,t}{=}\mathbf{p}_{ji,t}$ by construction; we choose order-invariance because directionality (``$i$ chases $j$'' vs.\ ``$j$ chases $i$'') is recoverable from $\mathbf{O}_i$ and $\mathbf{O}_j$, while our pretraining targets are largely symmetric in $(i,j)$.

\paragraph{Self-supervised objectives.} The social encoder is pretrained with two pairwise losses. The primary loss is \emph{pairwise HoA prediction} --- the dyadic analog of the individual HoA loss --- predicting histograms of a pairwise feature vector capturing velocity differences, heading geometry, and other dyadic kinematics (Appendix~\ref{app:features}). The auxiliary loss is \emph{pairwise distance regression} on the 13 per-keypoint and centroid distances, anchoring the pair representation to physical proximity. Because input spans are masked during pretraining, it also acts as a masked reconstruction of inter-animal distances; it targets relative positions because the frozen individual encoder already captures absolute ones. The objective is
\begin{equation}
\mathcal{L}_{\text{soc}} = \mathcal{L}_{\text{HoA-pair}} + \mathcal{L}_{\text{dist}},
\end{equation}
where $\mathcal{L}_{\text{HoA-pair}}$ is a per-pair-timestep KL between predicted and target histograms and $\mathcal{L}_{\text{dist}}$ is per-pair-timestep MSE. Loss equations and validity-mask definition are in Appendix~\ref{app:soc-losses}.

\section{Experiments}
\label{sec:experiments}

We evaluate MICE under four protocols on six mouse-pose datasets: linear probing (\S\ref{sec:probe}), unsupervised behavioral segmentation (\S\ref{sec:segmentation}), cross-dataset transfer (\S\ref{sec:transfer}), and few-shot label efficiency (\S\ref{sec:fewshot}), supplemented by layer-wise probing (\S\ref{sec:layer}), a social-stage ablation (\S\ref{sec:mechanism}), and leave-one-dataset-out transfer (\S\ref{sec:lodo}). Variable-$N$ and embedding-variant ablations appear in Appendix~\ref{app:ablations}.

\paragraph{Datasets.} The six datasets span $N{\in}\{1,2,3,4\}$ mice and 7--27 native keypoints, all mapped to the universal 12-keypoint vocabulary (\S\ref{sec:universal}): MABe22~\citep{sun2023mabe22}, CalMS21~\citep{sun2021multi}, CRIM13~\citep{burgos2012social}, MouseLockbox~\citep{reiske2025mouse}, AdaptableSnail, and NiftyGoldfinch (the last two are codenamed mouse tracks from MABe 2025~\citep{mabe2025}). Per-dataset details are in Appendix~\ref{app:datasets}.

\paragraph{Protocol.} We pretrain MICE on all six datasets jointly, then evaluate frozen embeddings via linear probing per-dataset (an MLP on frozen features); for comparison we also report \emph{single-dataset} pretraining. The headline MICE configuration is \texttt{ind+soc\_cat}; other pooling combinations appear as ablations in Tables~\ref{tab:main} and~\ref{tab:embed_variant_main}. Probe details are in Appendix~\ref{app:probe}.

\paragraph{Baselines.} We compare to four methods (BAMS~\citep{azabou2023relax}, PointNet~\citep{qi2017pointnet}, TBERT~\citep{sun2023mabe22}, hBehaveMAE~\citep{stoffl2024elucidating}) and two MICE ablations: \textbf{Progressive} (individual encoder with fixed-$N_{\max}$ padding) isolates the variable-$N$ contribution; \textbf{Multiscale} (multi-resolution supervision without widening attention windows) isolates the widening-window design. All methods share the universal keypoint mapping and identical downstream probes. MICE is the smallest competitive model (583K encoder parameters, about $7{\times}$ fewer than hBehaveMAE; Table~\ref{tab:params}). Implementation details in Appendix~\ref{app:baselines}.

\input{tables/main_results}

\subsection{Linear-Probe Behavior Classification}
\label{sec:probe}

Table~\ref{tab:main} reports macro-F1 under two pretraining regimes: \emph{single} (pretrain and probe on the same dataset) and \emph{combined} (joint pretrain on all six). Three findings stand out.

\textbf{MICE matches or exceeds the strongest baseline on every dataset under both regimes.} The largest gaps appear on the datasets with the most mice per recording --- MABe22 ($N{=}3$) and AdaptableSnail ($N\in\{3,4\}$) --- where variable-$N$ handling and joint social/individual modeling matter most. Fixed-ordering baselines such as hBehaveMAE trail MICE by a wider margin on these multi-animal datasets than on dyadic ones, supporting the variable-$N$ argument in \S\ref{sec:universal}. Appendices~\ref{app:cr-vanilla} and~\ref{app:cr-supervised} compare MICE with a vanilla transformer and with a supervised reference.

\textbf{The social stage adds consistent value on multi-animal datasets.} MICE (\texttt{ind\_cat}) vs.\ MICE (\texttt{ind+soc\_cat}) isolates the social encoder's contribution, positive on every multi-animal dataset under combined pretraining. The Progressive ablation, which stacks all $N$ animals into a single fixed-size input, lags MICE (\texttt{ind\_cat}) on most multi-animal datasets, showing that handling each mouse separately matters.

\textbf{Combined pretraining helps on most datasets.} On the five datasets with shared multi-animal structure (MABe22, CalMS21, CRIM13, AdaptableSnail, NiftyGoldfinch), combined pretraining matches or exceeds single-dataset pretraining for all MICE variants --- direct evidence for the foundation-model framing. On MouseLockbox it slightly hurts (single-animal-only structure limits transfer). Combined-regime MICE remains the best method under combined pretraining.
 
\subsection{Unsupervised Behavioral Segmentation}
\label{sec:segmentation}

We test whether frozen embeddings recover behavior \emph{without labels} by clustering them with $K$-means and scoring the segmentation against frame-level annotations. Per-window embeddings are partitioned into $K{=}50$ clusters; consecutive frames sharing a cluster are grouped into segments and mapped to ground-truth classes by majority vote. Table~\ref{tab:seg} reports ARI, NMI, and Boundary-F1 across the four datasets with frame-level labels. MICE leads ARI and NMI by sizable margins ($\sim$10 and $\sim$8 points over the strongest baseline) and matches hBehaveMAE on Boundary-F1; the three clustering metrics confirm MICE's embeddings align with behavior categories best among all methods. Protocol details, per-dataset results, and a comparison with Keypoint-MoSeq~\citep{weinreb2024keypoint} are in Appendix~\ref{app:segmentation}.

\input{tables/segmentation}

\begin{figure}[t]
\centering
\includegraphics[width=\linewidth]{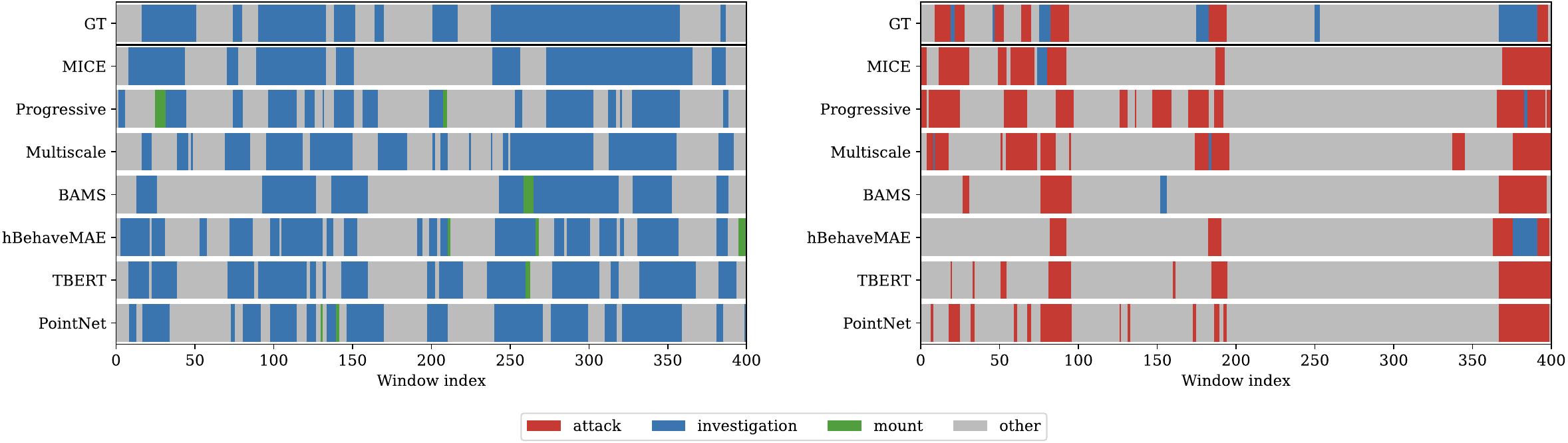}
\caption{Qualitative segmentation timelines from CalMS21, investigation episodes (left) and attack episodes (right). Each row shows one method's hard cluster assignments mapped to ground-truth classes. MICE matches ground-truth boundaries most closely; baselines either fragment episodes and produce false positives (e.g., Progressive and Multiscale) or miss real episodes (e.g., BAMS and hBehaveMAE on the early attack clusters).}
\label{fig:seg_timeline}
\end{figure}

\subsection{Cross-Dataset Transfer}
\label{sec:transfer}

Using the six single-dataset MICE checkpoints from \S\ref{sec:probe}, we linear-probe each on every dataset. As expected, the diagonal is best within each column among single-dataset checkpoints. Table~\ref{tab:transfer} reports the transfer matrix; the bottom row reproduces the combined-pretraining result. Two patterns emerge.

\textbf{Combined pretraining matches or exceeds single-dataset pretraining on five of six targets} --- direct evidence that joint pretraining transfers positive information across datasets. The exception is MouseLockbox, where combined pretraining trails by 2.3 points; we attribute this to label heterogeneity, since MouseLockbox's fine-grained puzzle-box behaviors differ from the social and locomotor labels dominating the other five datasets.

\textbf{Single-animal pretraining transfers poorly to multi-animal targets.} The MouseLockbox row produces the lowest off-diagonal scores on MABe22 and NiftyGoldfinch, matching the variable-$N$ argument from \S\ref{sec:universal}: a model never exposed to multi-animal interaction during pretraining cannot encode the social context downstream probes need. Appendix~\ref{app:cr-baseline-transfer} gives the same matrices for hBehaveMAE and PointNet.

\begin{table}[t]
\caption{Cross-dataset transfer: macro-F1 (\%) when MICE is pretrained on the row dataset and linear-probed on the column. Mean $\pm$ standard deviation over 5 random seeds. Diagonal cells correspond to single-regime MICE results in Table~\ref{tab:main} and the bottom row to the combined-regime results; MouseLockbox uses \texttt{ind\_cat}. \textbf{Bold} = best per column. AdaptSnail = AdaptableSnail; Goldfinch = NiftyGoldfinch; Lockbox = MouseLockbox.}
\label{tab:transfer}
\centering\footnotesize
\setlength{\tabcolsep}{3pt}
\begin{tabular}{l cccccc}
\toprule
& \multicolumn{6}{c}{Probe dataset} \\
\cmidrule(lr){2-7}
Pretrain dataset & MABe22 & CalMS21 & CRIM13 & AdaptSnail & Goldfinch & Lockbox \\
\midrule
MABe22       & $67.60_{\pm 0.45}$ & $72.65_{\pm 0.51}$ & $38.45_{\pm 0.78}$ & $20.85_{\pm 0.94}$ & $19.34_{\pm 0.71}$ & $13.41_{\pm 0.36}$ \\
CalMS21      & $58.74_{\pm 0.62}$ & $83.28_{\pm 0.39}$ & $38.34_{\pm 0.85}$ & $20.74_{\pm 0.97}$ & $26.59_{\pm 0.83}$ & $13.13_{\pm 0.41}$ \\
CRIM13       & $53.18_{\pm 0.71}$ & $72.05_{\pm 0.48}$ & $44.61_{\pm 0.72}$ & $20.82_{\pm 0.96}$ & $24.90_{\pm 0.81}$ & $15.55_{\pm 0.42}$ \\
AdaptSnail   & $62.45_{\pm 0.66}$ & $75.59_{\pm 0.52}$ & $27.64_{\pm 0.85}$ & $21.93_{\pm 1.04}$ & $22.17_{\pm 0.79}$ & $13.64_{\pm 0.39}$ \\
Goldfinch    & $47.93_{\pm 0.94}$ & $73.98_{\pm 0.55}$ & $32.91_{\pm 0.87}$ & $19.77_{\pm 0.91}$ & $31.08_{\pm 0.74}$ & $15.39_{\pm 0.45}$ \\
Lockbox      & $43.27_{\pm 1.08}$ & $74.25_{\pm 0.59}$ & $30.14_{\pm 0.92}$ & $20.75_{\pm 0.87}$ & $14.11_{\pm 0.62}$ & $\mathbf{15.93}_{\pm 0.43}$ \\
\midrule
\textbf{Combined} & $\mathbf{69.26}_{\pm 0.34}$ & $\mathbf{84.56}_{\pm 0.31}$ & $\mathbf{46.38}_{\pm 0.55}$ & $\mathbf{26.76}_{\pm 0.78}$ & $\mathbf{31.90}_{\pm 0.72}$ & $13.68_{\pm 0.37}$ \\
\bottomrule
\end{tabular}
\end{table}
 
\subsection{Few-Shot Label Efficiency}
\label{sec:fewshot}

We evaluate label efficiency on CalMS21 and CRIM13, two established mouse social-behavior benchmarks covering complementary difficulty regimes (CalMS21 moderately balanced; CRIM13 heavily imbalanced). We sweep the labeled training fraction across $\{1, 5, 10, 25, 50, 100\}\%$ and retrain only the MLP probe with the encoder frozen.

\begin{figure}[t]
\centering
\includegraphics[width=0.95\linewidth]{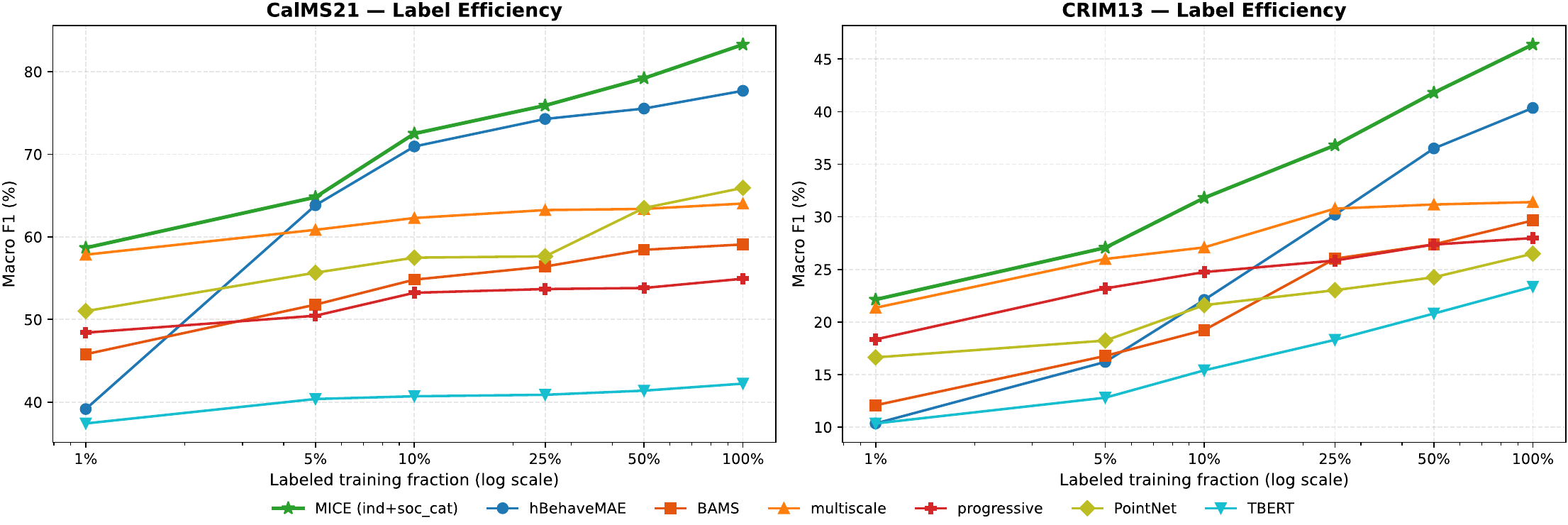}
\caption{Label efficiency on CalMS21 (left) and CRIM13 (right). Each method's frozen encoder feeds the same downstream MLP probe, retrained at six labeled-data fractions (log $x$-axis). Each curve is the mean macro-F1 over 5 random seeds on the official test split; standard deviations are within $\pm 1.5\%$ at every fraction (largest at the 1\% point, where training-set variance dominates) and are omitted from the plot for visual clarity. MICE leads at every label fraction; the gap over hBehaveMAE is largest at the lowest fractions.}
\label{fig:fewshot}
\end{figure}

MICE leads on both datasets at every label fraction (Figure~\ref{fig:fewshot}), and its margin over hBehaveMAE is largest at low fractions. With only 1\% of CalMS21 labels, MICE already matches BAMS's and exceeds TBERT's full-label performance, and substantially outperforms hBehaveMAE, whose representations degrade sharply when labels are scarce. With 5\% labels, MICE on CalMS21 nearly matches PointNet's full-label performance. A similar pattern holds on the harder CRIM13.

\subsection{Layer-wise Probing Analysis}
\label{sec:layer}

To localize \emph{where} discriminative information emerges, we extract embeddings at every intermediate layer with forward hooks and train an MLP probe per layer (encoder frozen). For MICE, this gives 16 probe points: 12 in the individual encoder and 4 in the social encoder. Probe-point details in Appendix~\ref{app:layerwise}. Figure~\ref{fig:layer} shows three patterns.

\begin{figure}[t]
\centering
\includegraphics[width=0.96\linewidth]{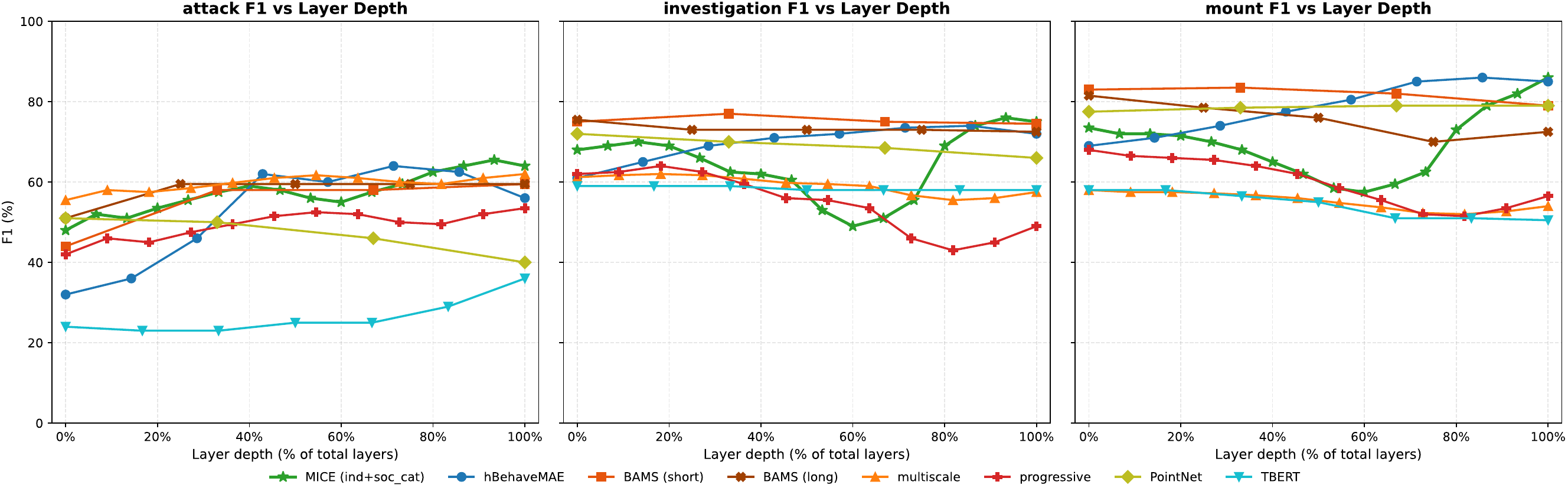}
\caption{Layer-wise linear-probe F1 on CalMS21 for three social behavior classes. $x$-axis: probe depth as percentage of total encoder depth. Each curve is the mean over 5 random seeds; per-layer standard deviations are within $\pm 1.0\%$ across all methods and are omitted from the plot for visual clarity. For MICE, $0$--$75\%$ is the individual encoder and $75$--$100\%$ is the social encoder. BAMS (short)/(long) = BAMS's two parallel branches; the long branch targets longer clips (Appendix~\ref{app:baselines}).}
\label{fig:layer}
\end{figure}

\textbf{Baselines show no depth-wise organization of behavior.} For TBERT, PointNet, and each of BAMS’s branches, early and late layers decode behavior about equally well, and no behavior shows up at a particular depth.

\textbf{MICE's individual encoder dips through Stage 3 on investigation and mount.} F1 dips from Stage 1 to 3 before recovering; Progressive shows the same pattern. The likely cause is what each stage represents: early layers (short attention windows) emphasize positional information, while later layers (long attention windows) emphasize velocity and motion integrated over time. Investigation and mount are pose-configurational (nose-to-flank, vertical posture), aligning with early-layer positional features; attack is motion-defined (lunging, fast directional change), aligning with later-layer velocity features and climbing gradually with depth.

\textbf{The social stage recovers MICE's F1 sharply; Progressive, lacking it, does not.} Starting at $75\%$ depth, MICE's F1 climbs steeply on investigation and mount, peaking near the final layer. Progressive dips at the same point but only partially recovers, ending well below MICE. This gap at $100\%$ is the empirical contribution of the Perceiver social encoder, enabled by combining each mouse's kinematic embedding (kept via a residual connection) with context about nearby animals.

\subsection{Social-Stage Ablation and Analysis}
\label{sec:mechanism}

\begin{table}[t]
\caption{Train-time social ablation: macro-F1 (\%) with the social stage retrained once per condition from one frozen individual encoder with correct, mis-paired, or removed partners, and probed likewise; mean $\pm$ standard deviation over 5 random probe seeds. First and last rows: combined-regime MICE (\texttt{ind+soc\_cat}, \texttt{ind\_cat}) from Table~\ref{tab:main}. MouseLockbox (one mouse, no social stage) is omitted. \textbf{Bold} = best per column.}
\label{tab:traintime}
\centering\footnotesize
\setlength{\tabcolsep}{3pt}
\begin{tabular}{l ccccc}
\toprule
Partners (train and probe) & MABe22 & CalMS21 & CRIM13 & AdaptSnail & Goldfinch \\
\midrule
correct (default) & $\mathbf{69.26}_{\pm 0.34}$ & $\mathbf{84.56}_{\pm 0.31}$ & $\mathbf{46.38}_{\pm 0.55}$ & $\mathbf{26.76}_{\pm 0.78}$ & $\mathbf{31.90}_{\pm 0.72}$ \\
mis-paired (shuffled) & $67.31_{\pm 0.54}$ & $81.19_{\pm 0.48}$ & $44.17_{\pm 0.71}$ & $22.79_{\pm 0.86}$ & $29.33_{\pm 0.91}$ \\
removed & $66.92_{\pm 0.50}$ & $81.00_{\pm 0.54}$ & $44.08_{\pm 0.66}$ & $21.83_{\pm 0.92}$ & $29.02_{\pm 0.83}$ \\
\midrule
\texttt{ind\_cat} only & $67.85_{\pm 0.42}$ & $81.22_{\pm 0.28}$ & $44.71_{\pm 0.47}$ & $22.66_{\pm 0.53}$ & $29.12_{\pm 0.48}$ \\
\bottomrule
\end{tabular}
\end{table}

To test whether the social stage's gain comes from modeling the other animals rather than from capacity, keypoint fusion, the residual path, or the pairwise objective, we retrain it with partners removed or mis-paired (real mice from other clips), keeping every parameter and loss, and probe each model under its training condition (Table~\ref{tab:traintime}). Two findings stand out.

\textbf{Without correct partner information, the social stage's gain disappears.} Both altered models fall back to about the level of the individual readout despite keeping the same capacity, keypoint fusion, and residual path; since mis-pairing keeps the pairwise objective non-trivial, extra supervision does not explain the gain either.

\textbf{The social embedding encodes how the animals are arranged relative to each other.} Linear probes show that it captures inter-animal distance and relative body orientation far better than the individual embedding, while adding nothing about the focal mouse's own movement (Appendix~\ref{app:cr-interp}): the social stage contributes relational information that per-animal kinematics cannot express. Direct animal-to-animal attention scores slightly lower on every dataset (Appendix~\ref{app:cr-traintime}) and, unlike the latents, scales with $N$.

\subsection{Leave-One-Dataset-Out Transfer}
\label{sec:lodo}

\begin{table}[t]
\caption{Leave-one-dataset-out transfer: macro-F1 (\%) when pretraining on all six datasets, the other five, or the target alone; mean $\pm$ standard deviation over 5 random probe seeds. Rows 1 and 3: combined and single-dataset MICE (Table~\ref{tab:transfer}); MouseLockbox uses \texttt{ind\_cat}. \textbf{Bold} = best per column.}
\label{tab:lodo}
\centering\footnotesize
\setlength{\tabcolsep}{3pt}
\begin{tabular}{l cccccc}
\toprule
Pretraining & MABe22 & CalMS21 & CRIM13 & AdaptSnail & Goldfinch & Lockbox \\
\midrule
all six datasets & $\mathbf{69.26}_{\pm 0.34}$ & $\mathbf{84.56}_{\pm 0.31}$ & $\mathbf{46.38}_{\pm 0.55}$ & $\mathbf{26.76}_{\pm 0.78}$ & $\mathbf{31.90}_{\pm 0.72}$ & $13.68_{\pm 0.37}$ \\
target left out (unseen) & $64.28_{\pm 0.52}$ & $84.05_{\pm 0.38}$ & $43.97_{\pm 0.72}$ & $20.45_{\pm 0.69}$ & $28.33_{\pm 0.82}$ & $12.92_{\pm 0.53}$ \\
target dataset only & $67.60_{\pm 0.45}$ & $83.28_{\pm 0.39}$ & $44.61_{\pm 0.72}$ & $21.93_{\pm 1.04}$ & $31.08_{\pm 0.74}$ & $\mathbf{15.93}_{\pm 0.43}$ \\
\bottomrule
\end{tabular}
\end{table}

We pretrain MICE on five datasets and probe it on the unseen sixth (Table~\ref{tab:lodo}). Two patterns emerge.

\textbf{MICE transfers to unseen datasets at a modest, dataset-dependent cost.} Leaving out CalMS21 or MouseLockbox barely matters, whereas leaving out AdaptableSnail or MABe22 costs more: transfer works best when other datasets cover the target's behaviors.

\textbf{A model that never saw the target nearly matches one pretrained on it.} Pretrained only on the other five datasets, MICE scores within 3.3 points of a model pretrained on the target alone, making a checkpoint pretrained on other datasets a reasonable starting point for a new dataset.

\section{Discussion and Conclusion}
\label{sec:discussion}

\paragraph{Why does the social channel complement the individual channel?} The individual and social pretraining objectives are non-overlapping (intra-animal kinematics vs.\ inter-animal pairwise quantities), so the two signals are not redundant: a fast-moving mouse next to another fast-moving mouse can be a chase or two parallel exploring mice, and only relative-heading disambiguates.

\paragraph{Limitations.} (i)~Synonym alignment is manually curated and must be extended by hand for new skeletons, though it needs no training data from a new lab. (ii)~Datasets are mouse-only and filmed from above; extending to other species would require re-curating the universal vocabulary. Bottom-up and 3D data remain untested. (iii)~The pretraining corpus, while spanning six datasets and $N\in\{1,2,3,4\}$, remains small relative to vision/language foundation models; scaling-law studies are left for future work. The range $N{\le}4$ is limited by available labeled datasets. Hyperparameters were not tuned beyond one loss-weight sweep (Appendix~\ref{app:cr-sweep}). (iv)~Behavior label sets are coarse; ethological subtleties below current annotation resolution will require new benchmarks.

\paragraph{Broader impact.} MICE could reduce annotation cost in behavioral neuroscience, accelerating drug discovery and welfare assessment. Adaptation for industrial-livestock surveillance carries dual-use risk; we recommend practitioners assess use-case ethics before deployment.

\paragraph{Conclusion.} MICE is a self-supervised foundation model for mouse social behavior combining a universal 12-keypoint vocabulary that aligns six heterogeneous datasets, a hierarchical individual encoder with linearly-widening attention windows, a Perceiver-based social encoder that natively handles variable animal counts, and pairwise pretraining objectives grounded in ethological features. Across the six datasets and four protocols, MICE achieves state-of-the-art results and combined multi-dataset pretraining matches or exceeds dataset-specific pretraining on most datasets --- evidence that a single checkpoint can serve diverse mouse-behavior tasks without architectural change.

\begin{ack}
This material is based upon work supported by the National Science Foundation under Award No.~\mbox{IIS-2336055}. Research reported in this publication was supported by the National Institute of Mental Health of the National Institutes of Health under award number RF1MH140429. Any opinions, findings, and conclusions or recommendations expressed in this material are those of the authors and do not necessarily reflect the views of the National Science Foundation. The content is solely the responsibility of the authors and does not necessarily represent the official views of the National Institutes of Health.
\end{ack}

{\small
\bibliographystyle{plainnat}
\bibliography{ref}
}

\newpage
\appendix

\section{Individual Encoder: Implementation Details}
\label{app:ind-schedule}

This section provides reproduction-level details for the individual encoder described in \S\ref{sec:ind}.

\paragraph{Tokenization.} The patching layer groups every $w{=}3$ consecutive frames at stride $s{=}1$ and projects each group via a single linear layer to a hidden state of dimension $D{=}32$, producing $T$ tokens per input pose sequence.

\paragraph{Within-stage layer iteration.} Denoting a single ALBERT block of stage $g$ as $f_g$ and the input to layer $\ell$ within stage $g$ as $\mathbf{h}^{(g)}_\ell$, the layer iteration is
\[
\mathbf{h}^{(g)}_{\ell+1} \;=\; f_g\!\left(\mathbf{h}^{(g)}_\ell;\, \omega(i)\right), \qquad \ell = 0, 1, \ldots, L_g - 1,
\]
where $\omega(i)$ is the attention window size at the global layer index $i = (g-1)\cdot L_g + \ell$. The input to stage $g{=}1$ is the patched token sequence; the input to stage $g{>}1$ is the output of the previous stage's last layer. Within-stage weight sharing means $f_g$ is the same block applied $L_g$ times.

\paragraph{HoA losses.}\label{app:ind-losses} The encoder predicts a soft histogram of the 55-dim kinematic feature vector $\boldsymbol{\phi}_t$ (Appendix~\ref{app:features}) over $B_{\text{HoA}}{=}32$ bins per feature. Predicting a soft histogram --- rather than the features themselves --- pushes the encoder toward short-window behavioral statistics rather than instantaneous values, closer to how ethologists characterize behavior. We apply HoA at two readout heads. The \emph{short/mid-scale head} reads off the concatenated output $\mathbf{H}_{\text{ind}}\in\mathbb{R}^{T\times 96}$ with histogram targets computed over a 7-frame window centered on each token; this supervises representations from earlier-stage layers (small attention windows). The \emph{long-scale head} reads off stage 3's output $\mathbf{h}^{(3)}\in\mathbb{R}^{T\times 32}$ with histogram targets computed over the full sequence of $T$ frames; this supervises the long-range representation alone. With $\mathcal{T}{=}\{1,\ldots,T\}$ indexing tokens, $p_{\text{pred}}^{s}, p_{\text{pred}}^{l}$ the predicted histograms at the two readouts, and $p_{\text{target}}^{s}, p_{\text{target}}^{l}$ the corresponding 7-frame and $T$-frame target histograms,
\begin{equation}
\mathcal{L}_{\text{HoA}}^{\text{short/mid}} = \frac{1}{|\mathcal{T}|} \sum_{t \in \mathcal{T}} \sum_{c=1}^{55} \mathrm{KL}\!\left(p_{\text{target}}^{s}(c, t)\,\big\|\,p_{\text{pred}}^{s}(c, t)\right),
\end{equation}
\begin{equation}
\mathcal{L}_{\text{HoA}}^{\text{long}} = \frac{1}{|\mathcal{T}|} \sum_{t \in \mathcal{T}} \sum_{c=1}^{55} \mathrm{KL}\!\left(p_{\text{target}}^{l}(c, t)\,\big\|\,p_{\text{pred}}^{l}(c, t)\right).
\end{equation}

\paragraph{Span-masked reconstruction loss.} Contiguous frame spans of length sampled uniformly from $[4,12]$ frames are masked in the input by zero-filling them in input keypoint space before patching. A linear layer then reads the predicted keypoint configuration $\hat{\mathbf{x}}_t \in \mathbb{R}^{K\times 2}$ off $\mathbf{H}_{\text{ind}}$ at each token's center frame:
\begin{equation}
\mathcal{L}_{\text{recon}} = \frac{1}{|\mathcal{T}|} \sum_{t \in \mathcal{T}} \big\|\hat{\mathbf{x}}_t - \mathbf{x}_t\big\|_2^2,
\end{equation}
where $\mathbf{x}_t$ is the ground-truth keypoint configuration. Span-masking (rather than per-frame masking) prevents the model from solving the task by linearly interpolating between adjacent unmasked frames: mouse motion is locally smooth at single-frame scales, but multi-frame spans cover direction changes that linear interpolation cannot recover.

\paragraph{Total objective.} $\mathcal{L}_{\text{ind}} = \mathcal{L}_{\text{HoA}}^{\text{short/mid}} + \mathcal{L}_{\text{HoA}}^{\text{long}} + \mathcal{L}_{\text{recon}}$, computed at all token positions for denser supervision than masked-only objectives.

\section{Social Encoder: Implementation Details}
\label{app:soc-equations}

This section provides reproduction-level details for the social encoder described in \S\ref{sec:soc}.

\paragraph{Input projection.} The frozen individual encoder produces per-animal hidden representations $\mathbf{H}_{\text{ind}}^{(n)}\in\mathbb{R}^{T\times D_{\text{ind}}}$ for $n=1,\ldots,N$. A linear projection fuses each per-animal embedding with the raw keypoints at the corresponding token-center frame:
\[
\mathbf{Z}_{n,t} \;=\; \mathrm{Linear}\!\left([\,\mathbf{H}_{\text{ind}}^{(n)}[t];\; \mathbf{x}_{n,t_c}\,]\right) \;\in\; \mathbb{R}^{d_{\text{model}}}, \qquad d_{\text{model}} = D_{\text{ind}} = 96,
\]
where $t_c$ is the center frame of token $t$ (the middle of the $w{=}3$-frame patching window) and $[\cdot;\cdot]$ denotes concatenation. The projection is a single linear layer $\mathbb{R}^{D_{\text{ind}} + 2K} \to \mathbb{R}^{d_{\text{model}}}$ with no nonlinearity.

\paragraph{Process operation details.} $\mathrm{SelfAttn}$ in Equation~\ref{eq:process} denotes $L_{\text{soc}}{=}2$ stacked self-attention layers applied to the latent codebook within each timestep; $\mathrm{TempAttn}$ applies temporal attention across the $T$ time tokens independently for each of the $M{=}16$ latent slots.

\paragraph{Pairwise HoA loss.}\label{app:soc-losses} For each unordered pair $(i,j)$ at each timestep $t$, the pairwise HoA head predicts a soft histogram of the 44-dim pairwise feature vector $\boldsymbol{\psi}_{ij}(t)$ (defined in Appendix~\ref{app:features}; $44 = 3K + 8$ for $K{=}12$, comprising per-keypoint distance and velocity-difference components plus eight relational features) over $B_{\text{HoA}}{=}32$ bins per feature. Let $\mathcal{P}$ denote the set of valid (pair, timestep) tuples, where validity requires both animals to have observed keypoints at $t$:
\begin{equation}
\mathcal{L}_{\text{HoA-pair}} = \frac{1}{|\mathcal{P}|} \sum_{(i,j,t) \in \mathcal{P}} \sum_{c=1}^{44} \mathrm{KL}\!\left(p_{\text{target}}(c, i, j, t)\,\big\|\,p_{\text{pred}}(c, i, j, t)\right),
\end{equation}
where $p_{\text{pred}}(c, i, j, t)$ is the predicted histogram for pair $(i,j)$ at timestep $t$ and feature channel $c$, and $p_{\text{target}}(c, i, j, t)$ is the target histogram computed from $\boldsymbol{\psi}_{ij}$ over a 7-frame window centered on $t$.

\paragraph{Pairwise distance loss.} The pairwise distance head reads off each pair representation $\mathbf{p}_{ij,t}$ to predict the 13-dim distance vector $\hat{\mathbf{d}}_{ij}(t)\in\mathbb{R}^{13}$ comprising the 12 per-keypoint Euclidean distances between animal $i$'s and animal $j$'s corresponding keypoints, plus the centroid-to-centroid distance:
\begin{equation}
\mathcal{L}_{\text{dist}} = \frac{1}{|\mathcal{P}|} \sum_{(i,j,t) \in \mathcal{P}} \big\|\hat{\mathbf{d}}_{ij}(t) - \mathbf{d}_{ij}(t)\big\|_2^2,
\end{equation}
where $\mathbf{d}_{ij}(t)\in\mathbb{R}^{13}$ is the corresponding ground-truth distance vector computed directly from the keypoint coordinates.

\paragraph{Validity mask.} The set $\mathcal{P}$ excludes (pair, timestep) tuples where either animal has no valid keypoints at timestep $t$ (e.g., out-of-frame or tracking failure). For datasets where validity is not annotated per-frame, all (pair, timestep) tuples are treated as valid. Sequences with no valid pair at any timestep are removed, so every pretraining batch contains valid pairs.

\paragraph{Total objective.} $\mathcal{L}_{\text{soc}} = \mathcal{L}_{\text{HoA-pair}} + \mathcal{L}_{\text{dist}}$, computed at all valid (pair, timestep) tuples. The two terms are weighted equally; we did not tune this weighting, and unit weighting keeps the objective interpretable.

\section{Self-Supervision Feature Definitions}
\label{app:features}

\paragraph{Individual-stage feature vector $\boldsymbol{\phi}_t$ (55-dim).} For each token $t$, the kinematic feature vector consumed by the HoA loss comprises:
\begin{itemize}
    \item \textbf{Per-keypoint velocity ($2K{=}24$ dims):} $x$- and $y$-velocity for each of the $K{=}12$ keypoints.
    \item \textbf{Per-keypoint centroid-relative position ($2K{=}24$ dims):} the offset $(x_k - x_{\text{ctr}}, y_k - y_{\text{ctr}})$ of each keypoint from the body centroid.
    \item \textbf{Body-length deviation (1 dim):} current body length minus the sequence-mean body length.
    \item \textbf{Forepaw--ear distance (1 dim):} a proxy for grooming, which involves the forepaws being brought near the head.
    \item \textbf{Heading direction (2 dims):} $(\cos h_t, \sin h_t)$ where $h_t$ is the angle of the tail-base--to--nose vector.
    \item \textbf{Turn direction (2 dims):} $(\cos \dot h_t, \sin \dot h_t)$ where $\dot h_t$ is the angular velocity of the heading.
    \item \textbf{Tail extension (1 dim):} a proxy for arousal, computed as the tail-base--to--tail-tip distance.
\end{itemize}
Total: $24 + 24 + 1 + 1 + 2 + 2 + 1 = 55$ dimensions. All features are normalized by sequence-mean body length for scale invariance. Histogram targets use $B_{\text{HoA}}{=}32$ bins per feature, computed over a 7-frame sliding window for the short/mid-scale head and over the full sequence for the long-scale head (Appendix~\ref{app:ind-schedule}).

\paragraph{Pairwise feature vector $\boldsymbol{\psi}_{ij}(t)$ (44-dim).} For each unordered pair of valid animals $(i,j)$ at token $t$:
\begin{itemize}
    \item \textbf{Per-keypoint inter-animal distance ($K{=}12$ dims):} $\|\mathbf{x}^{(k)}_i - \mathbf{x}^{(k)}_j\|_2 / \bar{\ell}$ for each keypoint $k$, normalized by the mean body length $\bar{\ell}$.
    \item \textbf{Per-keypoint velocity difference ($2K{=}24$ dims):} per-axis absolute velocity difference $|\mathbf{v}^{(k)}_{i,a} - \mathbf{v}^{(k)}_{j,a}|$ for each keypoint $k$ and axis $a \in \{x, y\}$, body-length normalized.
    \item \textbf{Centroid-centroid distance (1 dim):} $\|\mathbf{c}_i - \mathbf{c}_j\|_2 / \bar{\ell}$.
    \item \textbf{Heading similarity (1 dim):} $(1 + \cos(h_j - h_i)) / 2$, close to 1 when both animals face the same direction (parallel motion) and close to 0 when anti-parallel.
    \item \textbf{Heading orthogonality (1 dim):} $|\sin(h_j - h_i)|$, close to 1 when headings are perpendicular and close to 0 when collinear.
    \item \textbf{Face-to-face geometry score (1 dim):} 0 when one animal's nose points at the other's tail (chasing); 1 when noses point at each other (face-to-face confrontation).
    \item \textbf{Body-size asymmetry (1 dim):} $|\ell_i - \ell_j| / \max(\ell_i, \ell_j)$, indicating size differences relevant to dominance.
    \item \textbf{Forepaw--ear (grooming) asymmetry (1 dim):} the difference in forepaw--ear distance between the two animals.
    \item \textbf{Tail-extension (arousal) asymmetry (1 dim):} the difference in tail extension between the two animals.
    \item \textbf{Turn-rate similarity (1 dim):} angular distance between the two animals' instantaneous turning rates.
\end{itemize}
Total: $12 + 24 + 8 = 44$ dimensions, where the 12 + 24 = 36 per-keypoint dimensions plus 8 global-pair scalars give the $3K + 8$ form referenced in Appendix~\ref{app:soc-losses}.

\section{Dataset Details}
\label{app:datasets}

This section provides per-dataset specifics referenced from \S\ref{sec:experiments}. All datasets are top-down mouse-pose video keypoints; each native skeleton is mapped to the universal 12-keypoint vocabulary by named-synonym matching (\S\ref{sec:universal}), with missing slots zero-padded and masked from any feature that depends on them.

\paragraph{MABe22~\citep{sun2023mabe22}.} $N{=}3$ animals per sequence, 12 keypoints per mouse; its skeleton defines the universal vocabulary used by all other datasets. Behavior annotations are multi-label per window and cover a broad set of social behaviors.

\paragraph{CalMS21~\citep{sun2021multi}.} $N{=}2$ animals per sequence, 7 keypoints per mouse mapped to 7 of the 12 universal slots. Four mutually exclusive class labels: \emph{attack}, \emph{investigation}, \emph{mount}, \emph{other}. 70 training / 19 test sequences. Primary benchmark for dyadic social-interaction classification.

\paragraph{CRIM13~\citep{burgos2012social}.} $N{=}2$ animals per sequence, 7 keypoints per mouse mapped to 7 of the 12 universal slots. Eight mutually exclusive class labels: \emph{other}, \emph{approach}, \emph{attack}, \emph{disengage}, \emph{mount}, \emph{rear}, \emph{selfgroom}, \emph{sniff}, with the \emph{other} class dominating ($\sim$80\% of frames). Also used for the few-shot evaluation in \S\ref{sec:fewshot}.

\paragraph{AdaptableSnail (MABe 2025 challenge).} Variable group size, $N\in\{3,4\}$ mice per sequence, 10 keypoints per mouse mapped to 10 of the 12 universal slots. Used to evaluate the social encoder under within-dataset $N$ variation and to stress-test the universal mapping under a non-standard skeleton.

\paragraph{NiftyGoldfinch (MABe 2025 challenge).} $N{=}2$ mice per sequence. The set of keypoints differs between sequences (each file lists its own body-part names), so the mapping to the universal vocabulary is applied per sequence rather than once per dataset. This makes it the most stringent test of the universal mapping.

\paragraph{MouseLockbox~\citep{reiske2025mouse}.} $N{=}1$, 27 native DLC SuperAnimal keypoints mapped onto the 12 universal slots; the remaining 15 native keypoints are dropped. It covers fine-grained puzzle-box behaviors and is the only single-animal dataset in our evaluation.

\paragraph{Dataset sizes and sampling.} Table~\ref{tab:datasizes} lists training-split sizes. Batches are drawn from the concatenated corpus (proportional sampling, no reweighting). The corpus is imbalanced by $5.4{\times}$ (largest to smallest), not by orders of magnitude; for substantially more heterogeneous corpora we expect temperature-based sampling to become necessary.

\begin{table}[ht]
\caption{Training-split sizes (frames) and group sizes.}
\label{tab:datasizes}
\centering\footnotesize
\begin{tabular}{l r c}
\toprule
Dataset & Training frames & $N$ \\
\midrule
AdaptableSnail & 641{,}842 & 3--4 \\
CalMS21 & 507{,}738 & 2 \\
MouseLockbox & 427{,}380 & 1 \\
NiftyGoldfinch & 284{,}590 & 2 \\
CRIM13 & 223{,}362 & 2 \\
MABe22 & 118{,}560 & 3 \\
\bottomrule
\end{tabular}
\end{table}

\section{Linear Probe Details}
\label{app:probe}

For all probing experiments (``linear probing'' in the main text, following common usage for frozen-encoder evaluation), frozen embeddings are passed to a single-hidden-layer MLP with 256 ReLU units. Training uses Adam with weighted cross-entropy (inverse-frequency class weights), batch size 256, learning rate $1{\times}10^{-3}$, and 200 epochs with early stopping on macro-F1 over a held-out validation split. We report macro-F1 on the official test splits where available, otherwise on a 70/15/15 train/val/test split sampled at the recording level. Each result is the mean over 5 random seeds.

\paragraph{Embedding variants.} We evaluate six embedding variants of MICE: \texttt{ind\_mean}, \texttt{ind\_cat}, \texttt{soc\_mean}, \texttt{soc\_cat}, \texttt{ind+soc\_mean}, \texttt{ind+soc\_cat}. The prefix (\texttt{ind}, \texttt{soc}, or \texttt{ind+soc}) says which output is used: the individual encoder's, the social encoder's, or both concatenated. The suffix says how the $T{=}64$ tokens are combined: \texttt{mean} averages them into one 96- or 192-dim vector, and \texttt{cat} flattens them into a $T{\cdot}96$- or $T{\cdot}192$-dim vector. This is done per animal, and the vectors of all animals in the clip are then concatenated. The headline MICE result in Table~\ref{tab:main} uses \texttt{ind+soc\_cat}, and \texttt{ind\_cat} and \texttt{soc\_cat} appear in the same table as ablations; all six variants, including the mean-pooled ones, are compared on CalMS21 in Appendix~\ref{app:embed-variant}. Table~\ref{tab:readouts} summarizes them.

\begin{table}[ht]
\caption{MICE embedding readouts. $\mathbf{H}_{\text{ind}}$: concatenated individual-encoder output (Eq.~2); $\mathbf{O}$: per-animal social output (Eq.~\ref{eq:decode}); $T{=}64$ tokens. Each readout concatenates the embeddings of all animals in the clip, so \texttt{ind\_cat} already contains the partners' individual embeddings.}
\label{tab:readouts}
\centering\footnotesize
\begin{tabular}{l l l c}
\toprule
Readout & Source & Pooling over $T$ tokens & Dim.\ per animal \\
\midrule
\texttt{ind\_mean} / \texttt{ind\_cat} & $\mathbf{H}_{\text{ind}}\in\mathbb{R}^{T\times 96}$ & mean-pool / flatten & $96$ / $T{\cdot}96$ \\
\texttt{soc\_mean} / \texttt{soc\_cat} & $\mathbf{O}\in\mathbb{R}^{T\times 96}$ & mean-pool / flatten & $96$ / $T{\cdot}96$ \\
\texttt{ind+soc\_mean} / \texttt{ind+soc\_cat} & $[\mathbf{H}_{\text{ind}};\mathbf{O}]\in\mathbb{R}^{T\times 192}$ & mean-pool / flatten & $192$ / $T{\cdot}192$ \\
\bottomrule
\end{tabular}
\end{table}

\paragraph{The social readout also contains individual information.} The Decode step (Eq.~\ref{eq:decode}) adds the focal mouse's own projected embedding back in, so \texttt{soc} is not purely social. This is why \texttt{soc\_cat} alone is usually only modestly better than \texttt{ind\_cat}, and why \texttt{ind+soc\_cat}, which also keeps an unmixed copy of the individual embedding, does best.

\section{Unsupervised Segmentation: Protocol and Qualitative Details}
\label{app:segmentation}

\paragraph{Protocol.} For each method, we extract per-window embeddings on the test sequences with the most balanced mix of behavior labels. We run $K$-means with $K{=}50$ on the embeddings to produce hard cluster assignments. Consecutive frames with the same cluster assignment form a segment; each segment is mapped to the ground-truth class that appears most often in its frames (majority vote). ARI and NMI are computed at the frame level between predicted and ground-truth class labels. Boundary-F1 treats a predicted boundary as correct if it lies within 2 windows of any ground-truth boundary. All results are averaged over 5 seeds varying $K$-means initialization and the train/val/test split.

\paragraph{Qualitative observations from Figure~\ref{fig:seg_timeline}.} The figure shows two test sequences from CalMS21. In the left panel (investigation-rich), ground truth contains a long sustained investigation block at windows ${\sim}240$--$360$. MICE recovers most of it as two long segments separated by a short gap, with outer boundaries close to ground truth, while Progressive fragments the same block into several shorter segments. In the right panel (attack-rich), ground truth contains several attack clusters near the start (windows ${\sim}10$--$95$), one near window 190, and one near the end. MICE marks attack in both regions, whereas BAMS and hBehaveMAE miss most of the early clusters. Progressive and Multiscale also predict attack in windows where ground truth has none (e.g., ${\sim}120$--$180$ for Progressive), consistent with their tendency toward false positives.

\paragraph{Per-dataset results.} Table~\ref{tab:seg-perdataset} breaks the averages of Table~\ref{tab:seg} down by dataset.
MICE has the highest ARI and NMI on every dataset; on Boundary-F1 it leads on CalMS21, while hBehaveMAE leads on CRIM13 and NiftyGoldfinch and Progressive on AdaptableSnail.

\begin{table}[ht]
\caption{Per-dataset unsupervised segmentation results (\%, mean over 5 random seeds). The Avg.\ column is the average reported in Table~\ref{tab:seg}.}
\label{tab:seg-perdataset}
\centering\footnotesize
\setlength{\tabcolsep}{4pt}
\begin{tabular}{l ccccc}
\toprule
Model & CalMS21 & CRIM13 & AdaptSnail & Goldfinch & Avg. \\
\midrule
\multicolumn{6}{l}{\emph{ARI}} \\
PointNet & 57.90 & 51.49 & 10.74 & 14.77 & 33.73 \\
TBERT & 42.18 & 39.19 & 9.51 & 7.48 & 24.59 \\
BAMS & 16.54 & 13.81 & 5.04 & 1.71 & 9.28 \\
hBehaveMAE & 51.19 & 50.15 & 12.21 & 11.72 & 31.32 \\
Progressive (ours, ablation) & 11.78 & 13.76 & 8.85 & 2.92 & 9.33 \\
Multiscale (ours, ablation) & 40.39 & 43.32 & 11.32 & 9.48 & 26.13 \\
\textbf{MICE} (\texttt{ind+soc\_cat}) & 72.31 & 58.59 & 19.82 & 22.60 & 43.33 \\
\midrule
\multicolumn{6}{l}{\emph{NMI}} \\
PointNet & 44.24 & 40.02 & 10.68 & 15.07 & 27.50 \\
TBERT & 29.60 & 27.61 & 9.44 & 7.92 & 18.64 \\
BAMS & 11.78 & 10.83 & 6.22 & 3.27 & 8.03 \\
hBehaveMAE & 38.43 & 38.52 & 12.81 & 12.00 & 25.44 \\
Progressive (ours, ablation) & 9.14 & 10.59 & 10.03 & 2.76 & 8.13 \\
Multiscale (ours, ablation) & 32.61 & 31.35 & 12.54 & 9.12 & 21.41 \\
\textbf{MICE} (\texttt{ind+soc\_cat}) & 58.64 & 49.45 & 14.92 & 19.51 & 35.63 \\
\midrule
\multicolumn{6}{l}{\emph{Boundary-F1}} \\
PointNet & 31.98 & 27.83 & 9.26 & 35.70 & 26.19 \\
TBERT & 27.48 & 32.37 & 16.35 & 40.04 & 29.06 \\
BAMS & 15.96 & 16.77 & 16.11 & 16.95 & 16.45 \\
hBehaveMAE & 33.82 & 37.08 & 8.29 & 46.30 & 31.37 \\
Progressive (ours, ablation) & 17.43 & 19.69 & 16.67 & 41.94 & 23.93 \\
Multiscale (ours, ablation) & 28.72 & 34.23 & 13.93 & 44.43 & 30.33 \\
\textbf{MICE} (\texttt{ind+soc\_cat}) & 40.69 & 34.97 & 14.20 & 36.42 & 31.57 \\
\bottomrule
\end{tabular}
\end{table}

\paragraph{Comparison with Keypoint-MoSeq.} Keypoint-MoSeq (KPMS)~\citep{weinreb2024keypoint} is a widely used unsupervised method that segments pose into behavioral ``syllables''. We run it as the package intends for multi-animal data: each mouse is passed in as its own pose sequence with the same skeleton, and one model is fit over all of them, so KPMS learns a single syllable vocabulary and returns one syllable sequence per mouse. (Stacking the mice into one pose would place the body center between them, so the syllables would describe the pair's configuration rather than either animal.) We use two protocols. In \emph{train-fit}, one model is fit on all training recordings of a dataset and applied to each test recording, giving KPMS the same train/test separation as MICE. In \emph{per test recording}, a separate model is fit on each test recording; this is the easier setting for KPMS, since nothing has to transfer. Scoring follows our segmentation protocol: syllables are pooled into the same windows by majority vote, each syllable is mapped to the label it most often coincides with, and we report ARI, NMI and Boundary-F1 on CalMS21, CRIM13 and AdaptableSnail. Because KPMS returns one sequence per animal while the labels describe the whole scene, we score every animal and report the best one per recording, the most generous reading for KPMS (Table~\ref{tab:kpms}). KPMS has no term that couples animals: it assigns syllables from one egocentrically aligned pose at a time, whereas most labels in these datasets (e.g., attack, mount, sniff, approach) involve two animals.
MICE leads in all nine cells. Fitting KPMS on each test recording barely changes its ARI and NMI, so the gap is not due to transfer from training to test data; it is widest on CRIM13, the dataset with the most interaction-defined labels, where KPMS's ARI and NMI are close to zero.

\begin{table}[ht]
\caption{Unsupervised segmentation: Keypoint-MoSeq (KPMS) vs.\ MICE (\texttt{ind+soc\_cat}). Each cell gives ARI / NMI / Boundary-F1 (\%).}
\label{tab:kpms}
\centering\footnotesize
\begin{tabular}{l ccc}
\toprule
Dataset & KPMS (train-fit) & KPMS (per test recording) & MICE \\
\midrule
CalMS21 & 16.50 / 16.34 / 32.82 & 17.14 / 17.11 / 28.16 & 72.31 / 58.64 / 40.69 \\
CRIM13 & 0.75 / 0.91 / 0.27 & 0.74 / 1.45 / 1.95 & 58.59 / 49.45 / 34.97 \\
AdaptableSnail & 3.21 / 3.17 / 11.58 & 3.25 / 3.23 / 13.15 & 19.82 / 14.92 / 14.20 \\
\bottomrule
\end{tabular}
\end{table}

\section{Baseline Implementation Details}
\label{app:baselines}

All baselines share the universal keypoint mapping (\S\ref{sec:universal}) and raw-keypoint inputs, and feed their frozen embeddings to the same probe described in Appendix~\ref{app:probe}.

\paragraph{BAMS~\citep{azabou2023relax}.} Dual-branch dilated Conv1d (short-term + long-term) trained per-mouse with HoA prediction over a $K{\times}4$ velocity/shape feature target; mean-pooled across animals at inference. As in MICE's individual encoder, BAMS receives only raw keypoints as input, and the engineered features are used only as HoA prediction targets. This matches MICE's input and prevents the model from simply copying its targets from the input. Note that BAMS's long-term branch looks back about 1{,}365 frames, far longer than our 64-frame clips, so most of its input here is padding; its results should not be taken as representative of that branch.

\paragraph{PointNet~\citep{qi2017pointnet}.} Shared per-animal MLP with global max-pooling across the animal dimension yielding a permutation-invariant frame embedding; pretrained with temporal contrastive learning (positives = adjacent frames, negatives = different sequences).

\paragraph{TBERT~\citep{sun2023mabe22}.} BERT-style transformer over flattened multi-animal keypoints (fixed animal ordering), pretrained with masked-frame modeling ($\sim$40\% frame masking, MSE reconstruction).

\paragraph{hBehaveMAE~\citep{stoffl2024elucidating}.} Hierarchical masked autoencoder built on Hiera with $(2,1,n_{\text{kp}}{\times}2)$ patching, two pooling stages (4$\times$ temporal then animal-dimension pooling), three transformer stages (9 blocks total, embed dims 64/128/256), and multi-scale Conv3d fusion.

\paragraph{Progressive (ours, ablation).} The same architecture as MICE's individual encoder, but applied to all animals stacked into one multi-animal input with fixed $N_{\max}{=}4$ padding, and with no social stage; the multi-animal input makes it larger than MICE's per-animal individual encoder (Table~\ref{tab:params}). It isolates the contribution of per-animal encoding with a variable-$N$ social stage.

\paragraph{Multiscale (ours, ablation).} A single-transformer variant with simultaneous multi-resolution HoA heads at intermediate layers and no widening attention windows, in contrast to Progressive's staged hierarchy. It isolates MICE's staged, widening-window design from a flat multi-resolution alternative.

\section{Layer-wise Probing: Probe Locations}
\label{app:layerwise}

For the layer-wise probing analysis (\S\ref{sec:layer}), we probe MICE at 16 sequential depth points: 12 within the individual encoder and 4 within the social encoder. The mapping from probe index to architectural stage is:

\begin{center}
\footnotesize
\begin{tabular}{cll}
\toprule
Idx & Component & Operation  \\
\midrule
1--4   & Individual: Stage 1 & ALBERT layer iterations 1--4 (shared weights)  \\
5--8   & Individual: Stage 2 & ALBERT layer iterations 1--4 (shared weights)  \\
9--12  & Individual: Stage 3 & ALBERT layer iterations 1--4 (shared weights)  \\
\midrule
13     & Social: Encode & After cross-attention with $N$ animals + LayerNorm  \\
14     & Social: Process 1 & After first latent self-attention layer  \\
15     & Social: Process 2 & After second latent self-attention layer  \\
16     & Social: Temporal & After temporal attention across time tokens  \\
\bottomrule
\end{tabular}
\end{center}

To read out probe points 13--16, we pass the current latent state through the social encoder's $\mathrm{Decode}$ cross-attention, including its residual connection (\S\ref{sec:soc}). Each social probe therefore gives the per-animal output the encoder would produce if it stopped at that point. The input projection $\mathbf{Z}_{n,t}$ is not probed separately.

Baselines are probed at architecturally analogous points: BAMS's short and long branches are each probed at their successive layers and plotted as two separate curves; hBehaveMAE is probed at each of its hierarchical stage outputs (9 blocks across 3 stages); TBERT and PointNet are probed at all transformer/MLP layers respectively. To plot all methods on a common $x$-axis, each method's depth points are normalized to $\{0\%, \ldots, 100\%\}$ by total architectural depth.

\section{Ablations: Tables and Per-Ablation Interpretation}
\label{app:ablations}

This section contains two ablation studies: a variable-$N$ reorganization of the combined-regime linear-probe results, and an embedding-variant comparison on CalMS21.

\subsection{Variable-\texorpdfstring{$N$}{N} Analysis}
\label{app:varN}

Table~\ref{tab:varN} reorganizes Table~\ref{tab:main}'s combined-regime results by per-dataset $N$, with AdaptableSnail stratified by group size so each $N$ column reflects only recordings at that size. MICE wins every column; its lead over the baselines is negligible at $N{=}1$, modest at $N{=}2$, and substantial at $N{=}3$ and $N{=}4$. The most informative comparison is hBehaveMAE's reasonable $N{=}3$ performance on MABe22 versus its collapse on AdaptableSnail's $N{=}4$ recordings: the same architecture that handles fixed $N{=}3$ fails when $N$ varies within recordings. The MICE-family ablations sharpen the picture: the Progressive-vs-MICE gap is largest at $N{=}2$, where Progressive's fixed $N_{\max}{=}4$ padding leaves half the slots empty, and Multiscale performs comparably to Progressive overall. Because MICE's individual readout alone (\texttt{ind\_cat}) already outperforms both ablations in almost every column of Table~\ref{tab:main}, most of MICE's advantage comes from encoding each animal separately, with the social stage adding a further gain.

\begin{table}[ht]
\caption{Macro-F1 (\%) under combined pretraining, organized by per-dataset animal count. Mean $\pm$ standard deviation over 5 random seeds. Datasets per column: $N{=}1$: MouseLockbox; $N{=}2$: CalMS21, CRIM13, NiftyGoldfinch; $N{=}3$: MABe22 + AdaptableSnail-$N{=}3$; $N{=}4$: AdaptableSnail-$N{=}4$. For MICE, $N{=}1$ uses \texttt{ind\_cat} (social variants not applicable). \textbf{Bold} = best per column.}
\label{tab:varN}
\centering\small
\setlength{\tabcolsep}{4pt}
\begin{tabular}{l ccccc}
\toprule
& \multicolumn{4}{c}{Macro-F1 (\%) at dataset $N$} & Overall \\
\cmidrule(lr){2-5}
Model & $N{=}1$ & $N{=}2$ & $N{=}3$ & $N{=}4$ & avg \\
\midrule
TBERT \citep{sun2023mabe22}              & $13.55_{\pm 0.41}$ & $30.78_{\pm 0.58}$ & $23.62_{\pm 0.62}$ & $12.90_{\pm 0.69}$ & $20.21_{\pm 0.45}$ \\
PointNet \citep{qi2017pointnet}          & $13.42_{\pm 0.39}$ & $37.34_{\pm 0.51}$ & $37.40_{\pm 0.55}$ & $18.50_{\pm 0.62}$ & $26.67_{\pm 0.41}$ \\
BAMS \citep{azabou2023relax}             & $12.92_{\pm 0.40}$ & $35.37_{\pm 0.54}$ & $41.70_{\pm 0.52}$ & $17.90_{\pm 0.65}$ & $26.97_{\pm 0.43}$ \\
hBehaveMAE \citep{stoffl2024elucidating} & $13.21_{\pm 0.33}$ & $49.08_{\pm 0.45}$ & $41.98_{\pm 0.49}$ & $11.40_{\pm 0.58}$ & $28.92_{\pm 0.38}$ \\
\midrule
Progressive (ours, ablation)             & $13.54_{\pm 0.39}$ & $36.16_{\pm 0.49}$ & $55.72_{\pm 0.45}$ & $22.30_{\pm 0.61}$ & $31.93_{\pm 0.40}$ \\
Multiscale (ours, ablation)              & $13.04_{\pm 0.38}$ & $40.34_{\pm 0.46}$ & $56.72_{\pm 0.43}$ & $19.85_{\pm 0.59}$ & $32.49_{\pm 0.39}$ \\
\textbf{MICE}                            & $\mathbf{13.68}_{\pm 0.37}$ & $\mathbf{54.28}_{\pm 0.39}$ & $\mathbf{61.87}_{\pm 0.37}$ & $\mathbf{28.50}_{\pm 0.51}$ & $\mathbf{39.58}_{\pm 0.32}$ \\
\bottomrule
\end{tabular}
\end{table}

\paragraph{Progressive details.} Progressive has the same architecture as MICE's individual encoder but takes all animals stacked into one input tensor with $N_{\max}{=}4$ padding, and has no social stage. The Progressive-vs-MICE gap is largest at $N{=}2$ (18.1 points), where Progressive's fixed padding wastes capacity on padded slots ($N{=}2$ on a $N_{\max}{=}4$ tensor), and about 6 points at $N{=}3$ and $N{=}4$.

\paragraph{Multiscale details.} Multiscale is a variant of the individual encoder that supervises several temporal resolutions at once, without a staged hierarchy. It performs comparably to Progressive on most $N$ columns. Since both ablations trail MICE's individual readout alone in almost every column, the bulk of MICE's advantage comes from encoding each animal separately rather than from the staged hierarchy, with the social stage adding a further gain.

\paragraph{Per-animal pooling baselines.} BAMS and PointNet handle variable $N$ trivially via per-animal pooling but discard interaction structure entirely, reflected in scores that lag MICE by 10--25 points across $N{\geq}2$.

\paragraph{Single-dataset pretraining by group size.} Table~\ref{tab:varN-single} repeats this analysis with single-dataset pretraining, using the same grouping by $N$. The benefit of combined pretraining grows with the number of animals and is negative only at $N{=}1$, i.e., on MouseLockbox, where there is no social behavior for the other datasets to inform. Combined pretraining therefore helps most when the target involves the most social behavior.

\begin{table}[ht]
\caption{MICE macro-F1 (\%) by animal count under single-dataset vs.\ combined pretraining, grouped as in Table~\ref{tab:varN}. $\Delta$ = combined $-$ single.}
\label{tab:varN-single}
\centering\footnotesize
\begin{tabular}{c ccc}
\toprule
$N$ & Single-dataset & Combined & $\Delta$ \\
\midrule
1 & 15.93 & 13.68 & $-2.25$ \\
2 & 52.99 & 54.28 & $+1.29$ \\
3 & 56.77 & 61.87 & $+5.10$ \\
4 & 20.38 & 28.50 & $+8.12$ \\
\bottomrule
\end{tabular}
\end{table}

\paragraph{Foundation-model framing.} The 10.7-point gap on the overall average between MICE (39.58) and the strongest baseline (hBehaveMAE, 28.92) suggests that the benefits of handling variable $N$ and of jointly modeling individual and social behavior add up across diverse datasets. This is exactly the setting a foundation model is meant for: serving multi-animal datasets with different group sizes from a single checkpoint without retraining.

\subsection{Embedding Variant Analysis}
\label{app:embed-variant}

Table~\ref{tab:embed_variant_main} compares six embedding variants on CalMS21. Two patterns emerge. First, concatenation consistently beats mean-pooling, most of all for the individual channel, where averaging discards the temporal structure the hierarchical encoder is designed to capture. Second, combining the individual and social channels gives a small but consistent gain over either alone, indicating that they carry complementary information.

\begin{table}[ht]
\caption{Embedding-variant comparison on CalMS21 (combined-regime macro-F1, \%). Mean $\pm$ standard deviation over 5 random seeds. Channel: \texttt{ind} / \texttt{soc} / \texttt{ind+soc} = individual / social / both concatenated. Pooling: \texttt{mean} / \texttt{cat} = average / concatenation over $T$. \textbf{Bold} = best.}
\label{tab:embed_variant_main}
\centering\footnotesize
\setlength{\tabcolsep}{4pt}
\begin{tabular}{lcccccc}
\toprule
Variant & \texttt{ind\_mean} & \texttt{ind\_cat} & \texttt{soc\_mean} & \texttt{soc\_cat} & \texttt{ind+soc\_mean} & \texttt{ind+soc\_cat} \\
\midrule
F1 (\%) & $57.79_{\pm 0.42}$ & $81.22_{\pm 0.28}$ & $70.08_{\pm 0.36}$ & $81.86_{\pm 0.28}$ & $73.38_{\pm 0.34}$ & $\mathbf{84.56}_{\pm 0.31}$ \\
\bottomrule
\end{tabular}
\end{table}

\paragraph{Embedding-variant details.} The largest mean-vs-cat gap is on the individual channel (\texttt{ind\_mean}: 57.79 vs.\ \texttt{ind\_cat}: 81.22, a 23-point difference), where mean-pooling collapses the temporal structure most severely. The smaller gaps for the social and combined channels (12 and 11 points respectively) suggest that the social-stage representations are somewhat more robust to mean-pooling than the individual-stage representations, possibly because the social stage's temporal attention already aggregates information over time before pooling. The lift from \texttt{soc\_cat} (81.86) to \texttt{ind+soc\_cat} (84.56) is 2.7 points; \texttt{ind\_cat} alone is 81.22, so the ind+soc combination beats either channel alone but not by much. When embedding size is a constraint, \texttt{soc\_cat} alone is a reasonable choice at a cost of 2.7 points.

\section{Additional Analyses}
\label{app:cr}

These analyses complement \S\ref{sec:mechanism}--\ref{sec:lodo}.

\subsection{Retraining the social stage with different partner information}
\label{app:cr-traintime}

\paragraph{Design.} All runs start from the same frozen individual encoder. Because this encoder sees one animal at a time, it carries no partner information, and the individual readout (\texttt{ind\_cat}) is the same in every run. The social stage is then retrained from scratch for 500 epochs under one of three conditions:
\begin{itemize}
\item \emph{Correct partners}: the default setting.
\item \emph{Mis-paired partners}: each partner is replaced by the mouse in the same slot of another clip in the batch. Batches are grouped by animal count, so partners are real mice and only the pairing is wrong.
\item \emph{No partners}: partner slots are set to zero.
\end{itemize}
Only the social encoder's input is changed. All runs are compared at epoch 500. For each run, we train a fresh probe and test it under the same partner condition used in training, so no model sees unfamiliar input. With mis-paired partners, the pairwise loss remains meaningful, because its targets are computed from a real (if wrong) mouse. With no partners, these targets become trivial, so that run shows what happens without any partner information.

\paragraph{Direct animal-to-animal attention.} To test what the latent codebook adds over the simplest alternative, we replace the Perceiver latents with direct attention between animals, keeping correct partners and everything else fixed (Table~\ref{tab:nton}). Direct attention scores slightly lower than the Perceiver on all five datasets, by 0.7 to 1.8 points, so at these group sizes the latent codebook is at least as effective as the simplest alternative while also keeping cost independent of $N$.

\begin{table}[ht]
\caption{Perceiver latents vs.\ direct animal-to-animal attention in the social stage: macro-F1 (\%, \texttt{ind+soc\_cat}). Mean $\pm$ standard deviation over 5 random probe seeds. The Perceiver row is the combined-regime MICE result of Table~\ref{tab:main}. \textbf{Bold} = best per column.}
\label{tab:nton}
\centering\footnotesize
\setlength{\tabcolsep}{4pt}
\begin{tabular}{l ccccc}
\toprule
Social stage & MABe22 & CalMS21 & CRIM13 & AdaptSnail & Goldfinch \\
\midrule
Perceiver latents (MICE) & $\mathbf{69.26}_{\pm 0.34}$ & $\mathbf{84.56}_{\pm 0.31}$ & $\mathbf{46.38}_{\pm 0.55}$ & $\mathbf{26.76}_{\pm 0.78}$ & $\mathbf{31.90}_{\pm 0.72}$ \\
direct animal-to-animal attention & $68.13_{\pm 0.42}$ & $83.85_{\pm 0.40}$ & $45.11_{\pm 0.57}$ & $24.98_{\pm 0.85}$ & $30.92_{\pm 0.81}$ \\
\bottomrule
\end{tabular}
\end{table}

\subsection{Weighting the two individual-stage losses}
\label{app:cr-sweep}

\begin{table}[ht]
\caption{Individual-stage loss weights (macro-F1, \%, individual-only readout). $w_{\text{HoA}}$ and $w_{\text{rec}}$ weight the HoA and reconstruction losses; all models are pretrained on all six datasets, and the first row is the combined-regime \texttt{ind\_cat} result of Table~\ref{tab:main}. \textbf{Bold} = best per column.}
\label{tab:sweep}
\centering\footnotesize
\begin{tabular}{cc cc}
\toprule
$w_{\text{HoA}}$ & $w_{\text{rec}}$ & CalMS21 & CRIM13 \\
\midrule
1 & 1.00 & \textbf{81.22} & \textbf{44.71} \\
1 & 0.30 & 78.49 & 37.97 \\
1 & 0.10 & 76.46 & 36.01 \\
1 & 0.03 & 73.78 & 33.29 \\
1 & 0.00 & 74.10 & 33.92 \\
\midrule
0 & 1.00 & 78.84 & 34.02 \\
\bottomrule
\end{tabular}
\end{table}

Raising the reconstruction weight from 0.1 to 1 improves both datasets (CalMS21 76.46 to 81.22; CRIM13 36.01 to 44.71; Table~\ref{tab:sweep}), so we use equal weights. Removing the HoA loss ($w_{\text{HoA}}{=}0$) costs 10.7 F1 on CRIM13 and 2.4 on CalMS21.

\subsection{What the social stage encodes}
\label{app:cr-interp}

These analyses use the model pretrained on all six datasets, evaluated on the CalMS21 test set.

\paragraph{Relative position.} We fit linear models that predict simple measurements of the two mice from each frozen embedding (Table~\ref{tab:relational}). The social embedding predicts how the mice are positioned relative to each other far better than the individual embedding, and is no better at describing the focal mouse alone.

\begin{table}[ht]
\caption{How well a linear model predicts each measurement from the individual vs.\ social embedding ($R^2$; higher is better).}
\label{tab:relational}
\centering\footnotesize
\begin{tabular}{l ccc}
\toprule
Measurement & individual & social & difference \\
\midrule
Distance between the mice & 0.358 & 0.975 & $+0.617$ \\
Relative body orientation & 0.145 & 0.644 & $+0.499$ \\
Approach speed & $-0.221$ & 0.018 & $+0.239$ \\
Partner speed & $-0.059$ & $-0.032$ & $+0.027$ \\
\midrule
Focal mouse speed & 0.702 & 0.638 & $-0.064$ \\
Focal mouse body spread & 0.975 & 0.953 & $-0.022$ \\
\bottomrule
\end{tabular}
\end{table}

\subsection{Per-latent lift analysis}
\label{app:cr-lift}

We also examined whether individual latents respond selectively to particular behaviors.

\paragraph{Method.} For every token in the test set, we record the $M{=}16$ latents produced by the social encoder's Encode step, that is, after the latents have attended to the animals. For each latent, we rank all tokens by how strongly the latent is activated, take the top 100, and compute the class distribution among them. Dividing by the class prior gives a lift score, $P(c \mid \text{top 100 of latent } m) / P(c)$. A lift of 1 means the latent is active on that class no more often than chance; a lift of 10 means tokens where the latent is most active are ten times more likely to carry that label than a random token. No labels are used during pretraining; they are used only to read the latents afterwards.

\begin{table}[ht]
\caption{Most-enriched class per selected latent. Left: CalMS21 (priors: attack 4.97\%, investigation 23.4\%, mount 12.1\%, other 59.5\%). Right: CRIM13 (other prior 70.7\% among the analyzed tokens).}
\label{tab:lift}
\centering\footnotesize
\setlength{\tabcolsep}{3.5pt}
\begin{tabular}[t]{c l c c c}
\toprule
\multicolumn{5}{c}{CalMS21} \\
\midrule
Latent & Class & Lift & $P(c\mid\text{top})$ & $P(\text{other}\mid\text{top})$ \\
\midrule
3  & attack        & 10.86 & 54\% & 34\% \\
12 & attack        & 7.04  & 35\% & 49\% \\
14 & mount         & 3.05  & 37\% & 24\% \\
4  & investigation & 1.75  & 41\% & 59\% \\
\bottomrule
\end{tabular}
\hfill
\begin{tabular}[t]{c l c c c}
\toprule
\multicolumn{5}{c}{CRIM13} \\
\midrule
Latent & Class & Lift & $P(c\mid\text{top})$ & $P(\text{other}\mid\text{top})$ \\
\midrule
15 & disengage & 13.14 & 53\% & 36\% \\
12 & rear      & 9.85  & 18\% & 70\% \\
9  & attack    & 9.67  & 25\% & 51\% \\
1  & attack    & 7.35  & 19\% & 50\% \\
7  & disengage & 6.69  & 27\% & 56\% \\
0  & approach  & 5.29  & 19\% & 76\% \\
\bottomrule
\end{tabular}
\end{table}

\paragraph{Results.} On CalMS21, four of the sixteen latents are most enriched for attack, two of them strongly. Latent 3 is the clearest case: attack makes up 4.97\% of tokens overall but 54\% of the hundred tokens where latent 3 is most active. Eight latents are most enriched for other, at low lift (1.26--1.68). CRIM13 shows the same structure more sharply, with higher lift because its social classes are rarer: five latents are most enriched for attack and five for disengage, both interaction-defined classes. For CalMS21 latent 3 (attack, 54\% vs.\ 34\% other) and CRIM13 latent 15 (disengage, 53\% vs.\ 36\%), the named class is the outright majority of the top 100, and for CalMS21 latent 14 (mount, 37\% vs.\ 24\% other) it exceeds the background class.

\paragraph{Caveats.} Lift cannot exceed $1/P(c)$, so a common class can never reach a high lift: on CRIM13, the maximum lift is 1.41 for other (prior 70.7\%) but 38.5 for attack (prior 2.6\%). A high lift also does not mean the class dominates: other is still the largest class in the top 100 for twelve of sixteen CRIM13 latents. These latents therefore respond selectively to their class relative to how common it is, rather than responding only to it.

\subsection{Cross-dataset transfer for baselines}
\label{app:cr-baseline-transfer}

Table~\ref{tab:transfer} shows how MICE transfers across datasets but gives no reference for whether that behavior is specific to MICE. Tables~\ref{tab:transfer-hb} and~\ref{tab:transfer-pn} repeat the same protocol for hBehaveMAE, the strongest baseline, and PointNet: each model is pretrained on one dataset (rows) and probed on every dataset (columns).
Both baselines show the patterns we report for MICE. Within each column, pretraining on the target itself is the best single-dataset choice in 5 of 6 columns for hBehaveMAE and in all 6 for PointNet. Combined pretraining beats every single-dataset checkpoint on 5 of 6 targets for hBehaveMAE and 4 of 6 for PointNet, with MouseLockbox an exception for both. Pretraining on MouseLockbox gives the lowest off-diagonal score in 4 of the 5 multi-animal columns for hBehaveMAE and 2 of 5 for PointNet. Since these patterns appear across architectures that share only the data, they reflect the datasets rather than MICE.

\begin{table}[ht]
\caption{Cross-dataset transfer for hBehaveMAE: macro-F1 (\%) when pretrained on the row dataset and probed on the column dataset. Mean $\pm$ standard deviation over 5 random seeds. Diagonal cells and the bottom row are hBehaveMAE's single- and combined-regime results in Table~\ref{tab:main}.}
\label{tab:transfer-hb}
\centering\footnotesize
\setlength{\tabcolsep}{3pt}
\begin{tabular}{l cccccc}
\toprule
\diagbox{Pretrain}{Probe} & MABe22 & CalMS21 & CRIM13 & AdaptSnail & Goldfinch & Lockbox \\
\midrule
MABe22 & $42.72_{\pm 0.62}$ & $69.03_{\pm 0.49}$ & $35.22_{\pm 0.72}$ & $8.53_{\pm 0.64}$ & $20.36_{\pm 0.77}$ & $12.48_{\pm 0.56}$ \\
CalMS21 & $35.57_{\pm 0.74}$ & $76.05_{\pm 0.42}$ & $36.75_{\pm 0.78}$ & $8.11_{\pm 0.65}$ & $18.80_{\pm 0.75}$ & $13.67_{\pm 0.50}$ \\
CRIM13 & $32.93_{\pm 0.68}$ & $72.30_{\pm 0.44}$ & $38.30_{\pm 0.68}$ & $9.49_{\pm 0.70}$ & $19.20_{\pm 0.68}$ & $12.91_{\pm 0.51}$ \\
AdaptSnail & $34.26_{\pm 0.81}$ & $71.81_{\pm 0.58}$ & $26.91_{\pm 0.83}$ & $9.03_{\pm 0.53}$ & $18.05_{\pm 0.71}$ & $12.63_{\pm 0.63}$ \\
Goldfinch & $30.49_{\pm 0.66}$ & $72.54_{\pm 0.49}$ & $35.86_{\pm 0.76}$ & $8.66_{\pm 0.65}$ & $26.02_{\pm 0.59}$ & $14.35_{\pm 0.54}$ \\
Lockbox & $27.04_{\pm 0.73}$ & $64.33_{\pm 0.51}$ & $27.85_{\pm 0.80}$ & $7.73_{\pm 0.72}$ & $12.07_{\pm 0.69}$ & $14.58_{\pm 0.44}$ \\
\midrule
\textbf{Combined} & $47.82_{\pm 0.52}$ & $77.68_{\pm 0.34}$ & $40.35_{\pm 0.58}$ & $12.13_{\pm 0.53}$ & $29.21_{\pm 0.50}$ & $13.21_{\pm 0.33}$ \\
\bottomrule
\end{tabular}
\end{table}

\begin{table}[ht]
\caption{Cross-dataset transfer for PointNet, in the same format as Table~\ref{tab:transfer-hb}. Diagonal cells and the bottom row are PointNet's single- and combined-regime results in Table~\ref{tab:main}.}
\label{tab:transfer-pn}
\centering\footnotesize
\setlength{\tabcolsep}{3pt}
\begin{tabular}{l cccccc}
\toprule
\diagbox{Pretrain}{Probe} & MABe22 & CalMS21 & CRIM13 & AdaptSnail & Goldfinch & Lockbox \\
\midrule
MABe22 & $36.54_{\pm 0.71}$ & $58.36_{\pm 0.63}$ & $18.51_{\pm 0.76}$ & $7.59_{\pm 0.55}$ & $20.74_{\pm 0.63}$ & $12.90_{\pm 0.52}$ \\
CalMS21 & $30.25_{\pm 0.75}$ & $62.24_{\pm 0.54}$ & $17.93_{\pm 0.64}$ & $7.20_{\pm 0.46}$ & $19.10_{\pm 0.68}$ & $12.70_{\pm 0.49}$ \\
CRIM13 & $32.83_{\pm 0.68}$ & $56.13_{\pm 0.67}$ & $21.48_{\pm 0.65}$ & $6.72_{\pm 0.57}$ & $19.52_{\pm 0.55}$ & $13.64_{\pm 0.50}$ \\
AdaptSnail & $32.76_{\pm 0.72}$ & $61.50_{\pm 0.59}$ & $19.94_{\pm 0.69}$ & $7.91_{\pm 0.48}$ & $20.09_{\pm 0.62}$ & $13.00_{\pm 0.42}$ \\
Goldfinch & $28.38_{\pm 0.71}$ & $61.24_{\pm 0.59}$ & $20.48_{\pm 0.70}$ & $6.03_{\pm 0.61}$ & $21.73_{\pm 0.60}$ & $12.20_{\pm 0.56}$ \\
Lockbox & $26.59_{\pm 0.80}$ & $61.18_{\pm 0.65}$ & $18.93_{\pm 0.72}$ & $7.37_{\pm 0.53}$ & $17.06_{\pm 0.62}$ & $13.73_{\pm 0.42}$ \\
\midrule
\textbf{Combined} & $40.75_{\pm 0.64}$ & $65.93_{\pm 0.42}$ & $26.49_{\pm 0.57}$ & $19.59_{\pm 0.58}$ & $19.61_{\pm 0.52}$ & $13.42_{\pm 0.41}$ \\
\bottomrule
\end{tabular}
\end{table}

\subsection{Comparison with a vanilla transformer}
\label{app:cr-vanilla}

To test the individual encoder's design as a whole, we replace it with a vanilla variant: a single stage of 12 layers with full attention in every layer and a single readout at matched width, trained with the same setup as MICE. This removes both the widening attention windows and the staged readout. Because only the individual encoder differs, we compare the individual readout (\texttt{ind\_cat}), with MouseLockbox included (Table~\ref{tab:vanilla}). MICE's individual encoder outperforms the vanilla transformer on all six datasets, by 0.6 to 5.4 points, with the largest gaps on CRIM13 and NiftyGoldfinch, while also using fewer attention operations than full attention.

\begin{table}[ht]
\caption{MICE's individual encoder vs.\ a vanilla transformer (single stage, full attention): macro-F1 (\%, \texttt{ind\_cat}) under combined pretraining. Mean $\pm$ standard deviation over 5 random seeds. MICE values are from Table~\ref{tab:main}. Gap = MICE $-$ vanilla. \textbf{Bold} = best per row.}
\label{tab:vanilla}
\centering\footnotesize
\begin{tabular}{l ccc}
\toprule
Dataset & Vanilla transformer & MICE & Gap \\
\midrule
MABe22 & $65.32_{\pm 0.34}$ & $\mathbf{67.85}_{\pm 0.42}$ & $+2.53$ \\
CalMS21 & $80.16_{\pm 0.42}$ & $\mathbf{81.22}_{\pm 0.28}$ & $+1.06$ \\
CRIM13 & $39.33_{\pm 0.67}$ & $\mathbf{44.71}_{\pm 0.47}$ & $+5.38$ \\
AdaptableSnail & $21.30_{\pm 0.88}$ & $\mathbf{22.66}_{\pm 0.53}$ & $+1.36$ \\
NiftyGoldfinch & $23.23_{\pm 0.82}$ & $\mathbf{29.12}_{\pm 0.48}$ & $+5.89$ \\
MouseLockbox & $13.10_{\pm 0.52}$ & $\mathbf{13.68}_{\pm 0.37}$ & $+0.58$ \\
\bottomrule
\end{tabular}
\end{table}

\subsection{Supervised reference}
\label{app:cr-supervised}

As a reference point rather than a competitor, we take MICE's individual encoder with an identical configuration, initialize it randomly without self-supervised pretraining, attach a classification head, and train both end-to-end with cross-entropy on each dataset's behavior labels (Table~\ref{tab:supervised}). Unlike every method in Table~\ref{tab:main}, this model uses labels to train the encoder itself.
Supervised training is ahead on every dataset, as expected for a model that sees labels during training. The gap is widest on AdaptableSnail, where MICE's scores are weakest among the multi-animal datasets, and narrowest on CalMS21, where they are strongest.

\begin{table}[ht]
\caption{Supervised end-to-end training of MICE's individual encoder vs.\ self-supervised MICE (single-dataset regime, \texttt{ind+soc\_cat}; \texttt{ind\_cat} for MouseLockbox; Table~\ref{tab:transfer} diagonal): macro-F1 (\%). Mean $\pm$ standard deviation over 5 random seeds. Gap = supervised $-$ MICE.}
\label{tab:supervised}
\centering\footnotesize
\begin{tabular}{l ccc}
\toprule
Dataset & Supervised (end-to-end) & MICE & Gap \\
\midrule
MABe22 & $71.86_{\pm 0.62}$ & $67.60_{\pm 0.45}$ & $+4.26$ \\
CalMS21 & $85.86_{\pm 0.78}$ & $83.28_{\pm 0.39}$ & $+2.58$ \\
CRIM13 & $48.22_{\pm 0.74}$ & $44.61_{\pm 0.72}$ & $+3.61$ \\
AdaptableSnail & $29.87_{\pm 0.75}$ & $21.93_{\pm 1.04}$ & $+7.94$ \\
NiftyGoldfinch & $35.71_{\pm 0.66}$ & $31.08_{\pm 0.74}$ & $+4.63$ \\
MouseLockbox & $18.59_{\pm 0.63}$ & $15.93_{\pm 0.43}$ & $+2.66$ \\
\bottomrule
\end{tabular}
\end{table}

\subsection{Model sizes}
\label{app:cr-params}

\begin{table}[ht]
\caption{Encoder parameters used at inference (training-only heads excluded). MICE: 159{,}712 individual + 423{,}552 social = 583{,}264; with training-only heads, 772{,}053.}
\label{tab:params}
\centering\footnotesize
\begin{tabular}{l r r}
\toprule
Model & Parameters & vs.\ MICE \\
\midrule
PointNet & 28K & $0.05{\times}$ \\
\textbf{MICE} & \textbf{583K} & $1{\times}$ \\
Multiscale (ours, ablation) & 620K & $1.06{\times}$ \\
Progressive (ours, ablation) & 687K & $1.18{\times}$ \\
TBERT & 2.132M & $3.66{\times}$ \\
BAMS & 2.240M & $3.84{\times}$ \\
hBehaveMAE & 4.061M & $6.96{\times}$ \\
\bottomrule
\end{tabular}
\end{table}

MICE is small mainly because layers within each stage share weights (ALBERT-style).

\section{Training Hyperparameters}
\label{app:hyperparams}

\paragraph{Architecture.} \emph{Individual encoder}: patch size $w{=}3$, stride $s{=}1$; per-stage hidden dim $D{=}32$; three stages each with $L_g{=}4$ ALBERT-style layers and $h{=}8$ attention heads; concatenated output dim $D_{\text{ind}}{=}96$; attention windows follow the linear schedule of \S\ref{sec:ind} from $\omega_{\text{short}}{=}3$ to $\omega_{\text{long}}{=}T$. \emph{Social encoder}: $d_{\text{model}}{=}96$, learned latent codebook size $M{=}16$, social transformer layers $L_{\text{soc}}{=}2$, attention heads $h{=}8$. \emph{Pretraining heads}: HoA bins $B_{\text{HoA}}{=}32$ for both individual and pairwise heads. Sequence length $T{=}64$.

\paragraph{Span masking.} \emph{Individual encoder}: 25\% of frames per sequence are masked in contiguous spans of length sampled uniformly from $[4, 12]$ frames (zero-fill in input keypoint space). \emph{Social encoder (Masked Animal Modeling)}: 25\% of $T$ temporal tokens per animal are masked in contiguous spans of length sampled uniformly from $[2, 6]$ tokens, replaced by a learned mask token $\mathbf{m} \in \mathbb{R}^{d_{\text{model}}}$. Masking is applied only during training.

\paragraph{Loss weights.} The individual objective (Eq.~3) uses unit weights; equal weights worked best among those we tried (Appendix~\ref{app:cr-sweep}). The social objective also uses unit weights.

\paragraph{Optimization.} Adam optimizer with learning rate $1{\times}10^{-4}$, weight decay $1{\times}10^{-4}$, batch size 256. Mixed-precision (fp16) training with gradient scaling. Pretraining: 500 individual-stage epochs followed by 500 social-stage epochs with the individual encoder frozen. Each individual-stage epoch uses a random 5\% subsample of the pretraining corpus. Hardware: a single NVIDIA A40 GPU (48GB VRAM) per run; we ran many experiments in parallel rather than using multiple GPUs per run. The released code also includes a multi-GPU (DistributedDataParallel) training path.

\paragraph{Training and inference time.} On one NVIDIA A40, an individual-stage epoch takes about 149\,s and a social-stage epoch at most about 3.5\,min, so the 500 + 500-epoch schedule takes about 21\,h + 29\,h. Table~\ref{tab:timing} reports inference time for a batch of 32 clips of 64 frames, averaged over 200 timed iterations after 20 warm-up iterations. The individual encoder's cost does not depend on the number of animals, while the social stage adds about 1.6\,ms per additional animal.

\begin{table}[ht]
\caption{Inference time on one NVIDIA A40 (batch of 32 clips, 64 frames each).}
\label{tab:timing}
\centering\footnotesize
\begin{tabular}{c cccc}
\toprule
Animals & Individual encoder & Full model & Social stage & Full-model throughput \\
$N$ & (ms/batch) & (ms/batch) & (ms/batch) & (animal-sequences/s) \\
\midrule
1 & 7.07 & 12.49 & 5.43 & 2{,}562 \\
2 & 7.13 & 14.12 & 6.98 & 4{,}534 \\
3 & 7.12 & 15.72 & 8.60 & 6{,}109 \\
4 & 7.19 & 17.39 & 10.20 & 7{,}360 \\
\bottomrule
\end{tabular}
\end{table}

\paragraph{Tuning.} We did not tune architectural or training hyperparameters: the first configuration we tried was competitive, and we spent our compute on the method itself. Further tuning may improve results. How model size should grow with the amount of pretraining data is, in our view, the most important open question.

\end{document}

%% file: tables/main_results.tex
\begin{table}[t]
\caption{Linear-probing macro-F1 (\%) across the six datasets under two pretraining regimes: \emph{single} (pretrain and probe on the same dataset) and \emph{combined} (joint pretrain on all six, probe per dataset). Mean $\pm$ standard deviation over 5 random seeds. Within each regime, \textbf{bold} = best method and \underline{underline} = best baseline; $^\star$ and $^\dagger$ additionally mark these under the \emph{single} regime. ``---'': social variants do not apply to single-animal MouseLockbox. AdaptSnail = AdaptableSnail; Goldfinch = NiftyGoldfinch; Lockbox = MouseLockbox.}
\label{tab:main}
\centering\footnotesize
\setlength{\tabcolsep}{2pt}
\begin{tabular}{l l cccccc}
\toprule
Model & Regime & MABe22 & CalMS21 & CRIM13 & AdaptSnail & Goldfinch & Lockbox \\
\midrule
\multirow{2}{*}{PointNet \citep{qi2017pointnet}}
  & single   & $36.54_{\pm 0.7}$ & $62.24_{\pm 0.5}$ & $21.48_{\pm 0.7}$ & $7.91_{\pm 0.5}$ & $21.73_{\pm 0.6}$ & $13.73_{\pm 0.4}$ \\
  & combined & $40.75_{\pm 0.6}$ & $65.93_{\pm 0.4}$ & $26.49_{\pm 0.6}$ & $\underline{19.59}_{\pm 0.6}$ & $19.61_{\pm 0.5}$ & $13.42_{\pm 0.4}$ \\
\midrule
\multirow{2}{*}{TBERT \citep{sun2023mabe22}}
  & single   & $20.18_{\pm 0.8}$ & $42.23_{\pm 0.7}$ & $19.01_{\pm 0.7}$ & $\underline{11.12}^{\dagger}_{\pm 0.6}$ & $13.69_{\pm 0.6}$ & $14.20_{\pm 0.4}$ \\
  & combined & $25.45_{\pm 0.7}$ & $56.40_{\pm 0.5}$ & $23.35_{\pm 0.6}$ & $13.72_{\pm 0.5}$ & $12.59_{\pm 0.5}$ & $\underline{13.55}_{\pm 0.4}$ \\
\midrule
\multirow{2}{*}{BAMS \citep{azabou2023relax}}
  & single   & $\underline{43.90}^{\dagger}_{\pm 0.6}$ & $74.98_{\pm 0.4}$ & $\underline{41.41}^{\dagger}_{\pm 0.7}$ & $8.95_{\pm 0.5}$ & $14.86_{\pm 0.6}$ & $13.30_{\pm 0.4}$ \\
  & combined & $46.17_{\pm 0.5}$ & $59.08_{\pm 0.5}$ & $29.65_{\pm 0.6}$ & $18.67_{\pm 0.6}$ & $17.39_{\pm 0.5}$ & $12.92_{\pm 0.4}$ \\
\midrule
\multirow{2}{*}{hBehaveMAE \citep{stoffl2024elucidating}}
  & single   & $42.72_{\pm 0.6}$ & $\underline{76.05}^{\dagger}_{\pm 0.4}$ & $38.30_{\pm 0.7}$ & $9.03_{\pm 0.5}$ & $\underline{26.02}^{\dagger}_{\pm 0.6}$ & $\underline{14.58}^{\dagger}_{\pm 0.4}$ \\
  & combined & $\underline{47.82}_{\pm 0.5}$ & $\underline{77.68}_{\pm 0.3}$ & $\underline{40.35}_{\pm 0.6}$ & $12.13_{\pm 0.5}$ & $\underline{29.21}_{\pm 0.5}$ & $13.21_{\pm 0.3}$ \\
\midrule
\midrule
\multirow{2}{*}{Progressive (ours, ablation)}
  & single   & $53.93_{\pm 0.6}$ & $54.93_{\pm 0.6}$ & $18.71_{\pm 0.8}$ & $14.17_{\pm 0.6}$ & $9.50_{\pm 0.5}$ & $15.78_{\pm 0.4}$ \\
  & combined & $61.87_{\pm 0.4}$ & $66.97_{\pm 0.4}$ & $27.98_{\pm 0.6}$ & $23.63_{\pm 0.6}$ & $13.54_{\pm 0.5}$ & $13.54_{\pm 0.4}$ \\
\midrule
\multirow{2}{*}{Multiscale (ours, ablation)}
  & single   & $52.28_{\pm 0.6}$ & $64.03_{\pm 0.5}$ & $24.54_{\pm 0.7}$ & $14.61_{\pm 0.6}$ & $17.43_{\pm 0.5}$ & $15.72_{\pm 0.4}$ \\
  & combined & $63.68_{\pm 0.4}$ & $74.24_{\pm 0.4}$ & $31.40_{\pm 0.6}$ & $20.92_{\pm 0.6}$ & $15.39_{\pm 0.5}$ & $13.04_{\pm 0.4}$ \\
\midrule
\multirow{2}{*}{\textbf{MICE} (\texttt{ind\_cat})}
  & single   & $65.58_{\pm 0.5}$ & $79.76_{\pm 0.3}$ & $39.12_{\pm 0.6}$ & $20.91_{\pm 0.6}$ & $28.55_{\pm 0.5}$ & $\mathbf{15.93}^{\star}_{\pm 0.4}$ \\
  & combined & $67.85_{\pm 0.4}$ & $81.22_{\pm 0.3}$ & $44.71_{\pm 0.5}$ & $22.66_{\pm 0.5}$ & $29.12_{\pm 0.5}$ & $\mathbf{13.68}_{\pm 0.4}$ \\
\midrule
\multirow{2}{*}{\textbf{MICE} (\texttt{soc\_cat})}
  & single   & $67.13_{\pm 0.5}$ & $79.34_{\pm 0.3}$ & $40.92_{\pm 0.5}$ & $21.34_{\pm 0.5}$ & $29.42_{\pm 0.5}$ & --- \\
  & combined & $68.88_{\pm 0.4}$ & $81.86_{\pm 0.3}$ & $44.09_{\pm 0.5}$ & $25.15_{\pm 0.4}$ & $30.35_{\pm 0.6}$ & --- \\
\midrule
\multirow{2}{*}{\textbf{MICE} (\texttt{ind+soc\_cat})}
  & single   & $\mathbf{67.60}^{\star}_{\pm 0.5}$ & $\mathbf{83.28}^{\star}_{\pm 0.4}$ & $\mathbf{44.61}^{\star}_{\pm 0.7}$ & $\mathbf{21.93}^{\star}_{\pm 1.0}$ & $\mathbf{31.08}^{\star}_{\pm 0.7}$ & --- \\
  & combined & $\mathbf{69.26}_{\pm 0.3}$ & $\mathbf{84.56}_{\pm 0.3}$ & $\mathbf{46.38}_{\pm 0.6}$ & $\mathbf{26.76}_{\pm 0.8}$ & $\mathbf{31.90}_{\pm 0.7}$ & --- \\
\bottomrule
\end{tabular}
\end{table}

%% file: tables/segmentation.tex
\begin{table}[t]
\caption{Unsupervised behavioral segmentation via $K$-means clustering ($K{=}50$) on frozen embeddings, mean $\pm$ standard deviation across the four datasets with frame-level labels (CalMS21, CRIM13, AdaptableSnail, NiftyGoldfinch) over 5 random seeds. All metrics reported as \%. \textbf{Bold} = best, \underline{underline} = best baseline. MICE leads ARI and NMI by sizable margins and matches hBehaveMAE on Boundary-F1 (within seed variation).}
\label{tab:seg}
\centering\small
\setlength{\tabcolsep}{6pt}
\begin{tabular}{lccc}
\toprule
Model & ARI (\%) $\uparrow$ & NMI (\%) $\uparrow$ & Boundary-F1 (\%) $\uparrow$ \\
\midrule
PointNet \citep{qi2017pointnet}             & $\underline{33.73}_{\pm 0.24}$ & $\underline{27.50}_{\pm 0.18}$ & $26.19_{\pm 0.31}$ \\
TBERT \citep{sun2023mabe22}                 & $24.59_{\pm 0.27}$ & $18.64_{\pm 0.21}$ & $29.06_{\pm 0.28}$ \\
BAMS \citep{azabou2023relax}                & $9.28_{\pm 0.34}$  & $8.03_{\pm 0.28}$  & $16.45_{\pm 0.36}$ \\
hBehaveMAE \citep{stoffl2024elucidating}    & $31.32_{\pm 0.21}$ & $25.44_{\pm 0.16}$ & $\underline{31.37}_{\pm 0.24}$ \\
\midrule
Progressive (ours, ablation)                & $9.33_{\pm 0.32}$  & $8.13_{\pm 0.26}$  & $23.93_{\pm 0.29}$ \\
Multiscale (ours, ablation)                 & $26.13_{\pm 0.23}$ & $21.41_{\pm 0.19}$ & $30.33_{\pm 0.25}$ \\
\textbf{MICE} (\texttt{ind+soc\_cat})       & $\mathbf{43.33}_{\pm 0.17}$ & $\mathbf{35.63}_{\pm 0.13}$ & $\mathbf{31.57}_{\pm 0.21}$ \\
\bottomrule
\end{tabular}
\end{table}